\documentclass[twoside]{article}

\usepackage[accepted]{aistats2022}
\usepackage{xcolor}       
\usepackage{graphicx}
\usepackage{float}
\usepackage{amsmath,amssymb,amsfonts}
\usepackage{fancyhdr}

\usepackage{outlines}
\usepackage{enumitem}
\setenumerate[1]{label=\Alph*}
\setenumerate[2]{label*=\arabic*.}

\usepackage[colorlinks,citecolor=blue,urlcolor=blue,bookmarks=false,hypertexnames=true]{hyperref} 

\newtheorem{innercustomthm}{}
\newenvironment{customthm}[1]
  {\renewcommand\theinnercustomthm{#1}\innercustomthm}
  {\endinnercustomthm}

\newtheorem{theorem}{Theorem}[section]

\newtheorem{lemma}[theorem]{Lemma}

\usepackage[round]{natbib}

\begin{document}

%

%

\twocolumn[

\aistatstitle{Efficient Inference Without Trading-off Regret in Bandits:\\ An Allocation Probability Test for Thompson Sampling
}

\aistatsauthor{ Nina Deliu \And Joseph J. Williams \And  Sofia S. Villar}

\aistatsaddress{ MRC - Biostatistics Unit\\ University of Cambridge  \And  Department of Computer Science\\ University of Toronto \And MRC - Biostatistics Unit\\ University of Cambridge} ]

\begin{abstract}

Using bandit algorithms to conduct adaptive randomised experiments can minimise regret, but it poses major challenges for statistical inference (e.g., biased estimators, inflated type-I error and reduced power). Recent attempts to address these challenges typically impose restrictions on the exploitative nature of the bandit algorithm–trading off regret–and require large sample sizes to ensure asymptotic guarantees. However, large experiments generally follow a successful pilot study, which is tightly constrained in its size or duration. 
Increasing power in such small pilot experiments, without limiting the adaptive nature of the algorithm, can allow promising interventions to reach a larger experimental phase. In this work we introduce a novel hypothesis test, uniquely based on the allocation probabilities of the bandit algorithm, and without constraining its exploitative nature or requiring a minimum experimental size. We characterise our \textit{Allocation Probability Test} when applied to Thompson Sampling, presenting its asymptotic theoretical properties, and illustrating its finite-sample performances compared to state-of-the-art approaches. 
We demonstrate the regret and inferential advantages of our approach–particularly in small samples–in both extensive simulations and in a real-world experiment on mental health aspects.


\end{abstract}

\section{Introduction}

Over the last decade, a number of application domains, going from
online advertising~\citep{li2010contextual,chapelle2011empirical} to 
mobile health~\citep[e.g., mental health or physical activity;][]{aguilera2020mhealth, Figueroa10.1093/abm/kaab028}, have embraced and led the way in the use of \textit{multi-armed bandit} (MAB) algorithms to conduct adaptive experimentation. Such algorithms can substantially improve performance in terms of particular objectives, such as achieving optimal asymptotic regret bounds 
~\citep[see the seminal works of][]{auer2002finite,agrawal2013further}.

However, the increased use of 
MAB algorithms for conducting 
experiments, has been followed by the realisation that using adaptively-collected data for important secondary inferential objectives 
may be statistically challenging. Most specifically, \textbf{ traditional inference methods are typically not valid or reliable when used in MAB-collected data}. It is well recognized  that the sequential dependence induced by adaptive algorithms can lead to \emph{considerable biases} in classical estimators~\citep{villar2015multi,bowden2017unbiased,ShinNEURIPS2019_65b1e92c}, that persist even in the infinite data limit, with \emph{asymptotic distributions differing from the predicted central limit theorem (CLT) behavior}~\citep{deshpande2018accurate,hadad2021confidence}. More recently, issues have been detected in hypothesis testing as well~\citep{villar2015multi,rafferty2019statistical}, attracting a substantial attention within the machine learning (ML) community~\citep{Zhang_NEURIPS2020,yao_power,hadad2021confidence}. These issues include \emph{type-I error inflation} (i.e., higher than a pre-defined target $\alpha$) and \emph{reduced statistical power} to reject the null (or equivalently, increased \textit{type-II error}). 

In the last years, clinical trials (CTs) have seen an increased uptake of flexible, sequential designs to deliver efficiency or ethical gains~\citep{stallard2020efficient, shen2020learning}. This trend became evident for trials research during the COVID-19 crisis~\citep[see e.g.,][]{remap2021interleukin, johnson2021quantifying}.
Nevertheless, in CTs, the need for ensuring validity with adaptively-collected data~\citep{pallmann2018adaptive,jennison2013interim} has been long recognised and has traditionally dominated over other goals (including regret minimisation). 
Applied biostatistics has led the way in developing valid inference methods in a number of adaptive settings, but not for MAB algorithms. For example, the approach of~\cite{Zhang_NEURIPS2020}, which is robust even in case of non-stationarity, shares the rationale in~\cite{karrison2003group} and \cite{jennison2001group}, where a stratified analysis by batches is shown to eliminate bias due to time trends.

In this work, we focus on the problem of \emph{hypothesis testing} on data collected by regret-minimization MAB algorithms, such as Thompson Sampling (TS). We introduce a novel test statistic, named \textit{Allocation Probability Test} (AP-test), designed to overcome key limitations of existing testing approaches (e.g, imposing restrictions on the algorithm). We aim to offer a valid yet more practical solution to applied communities.

\begin{figure}[ht]
    \centering
    \includegraphics[scale = 0.44]{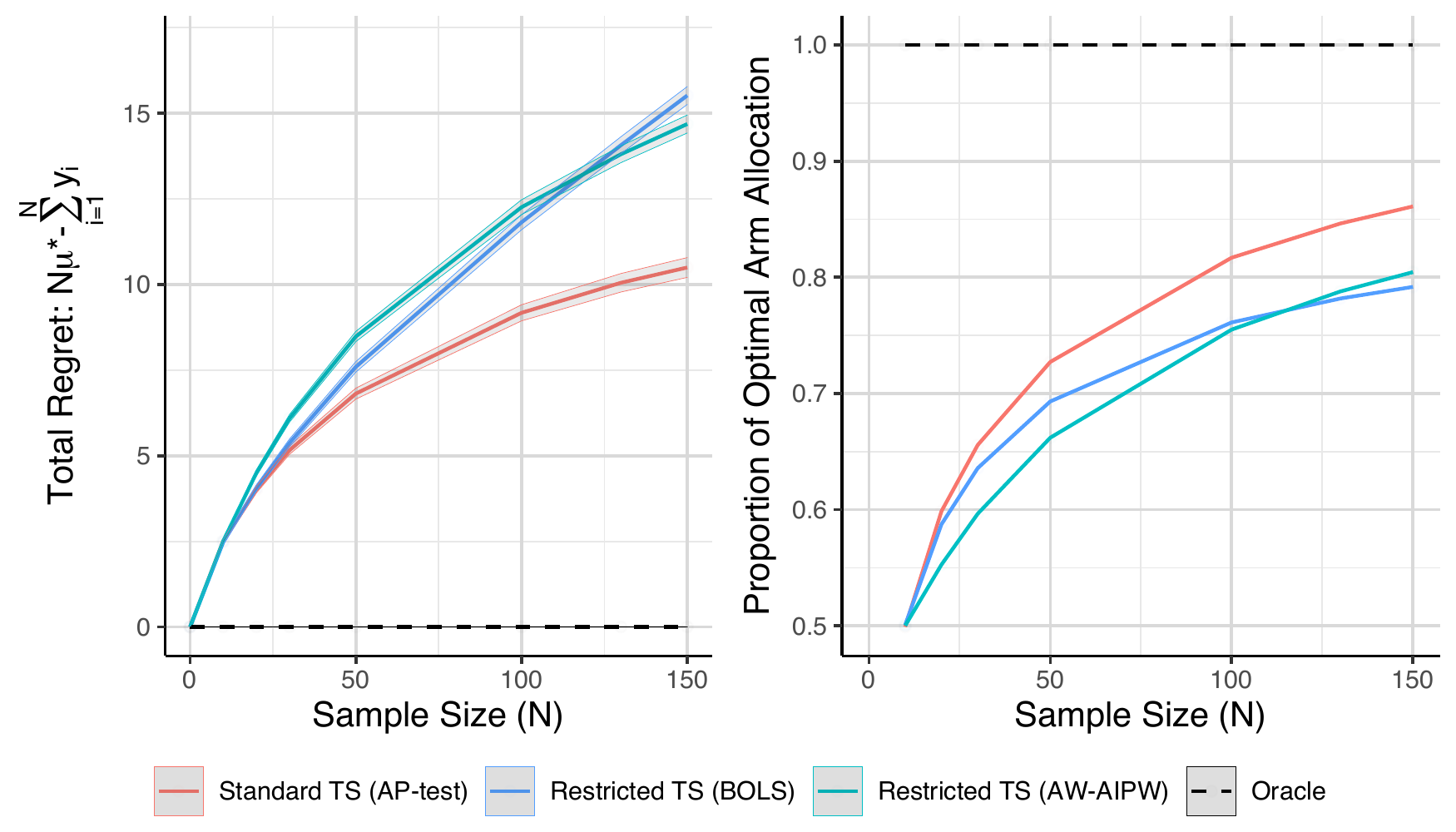}
    \caption{Regret and proportion of optimal arm allocation in a two-armed setting (i.e., arms $A_i \in \{0,1\}$), 
    with rewards 
    $Y_i (A_i = 0) \sim N(\mu_0 = 0, \sigma^2 = 1)$ and $Y_i (A_i = 1) \sim N(\mu_1 = \mu^* = 0.5, \sigma^2 = 1)$, $i=1,\dots,N$.}
    \label{fig:regretex}
\end{figure}

\textbf{Existing Works and Limitations.} Hypothesis testing for MAB-collected data has been very recently considered in the ML literature~\citep{Zhang_NEURIPS2020, hadad2021confidence}, with a strong focus on \emph{type-I error control}. The predominant approach has been to establish theoretical conditions under which traditional Z-tests can lead to valid \emph{asymptotic inference} (e.g., by restricting arms' allocation probabilities to lie within a range). Notably, while such constraints on the MAB algorithm may ensure CLT guarantees in adaptive settings, they may also considerably increase regret. 
To illustrate this, Figure~\ref{fig:regretex} displays the regret attainable by both standard TS (on which we build our AP-test), 
the restricted TS policies used in the \textit{Adaptively-Weighted Augmented Inverse Probability Weighing} \citep[AW-AIPW;][]{hadad2021confidence} and \textit{Batched Ordinary Least Squares}~\citep[BOLS;][]{Zhang_NEURIPS2020} methods, and an \textit{Oracle} that always allocates the highest mean reward arm~\citep{besbes2014stochastic}. Note that regret is the difference between the performance of the oracle and the TS policies.

Additionally, the above inference methods for bandit algorithms are asymptotically valid and typically require either a large experiment or a large batch size. As shown in Section~\ref{sec:results}, they may either fail to control type-I error (AW-AIPW in small samples), or have practically no power (i.e., $<0.1$ for BOLS in a batch size of $3$ and regardless the sample size). Note that the BOLS is applicable only for batch sizes $\geq 3$.  Existing valid inferential approaches for finite samples are also inefficient in terms of statistical power~\citep[see e.g., randomisation-based inference in][]{simon2011using,villar2018response}. Thus, \textbf{existing options for valid hypothesis testing for MAB collected data are applicable only to large experiments (and/or large batches), with a moderate reward performance and insufficient statistical power}.

\textbf{Practical Motivation.}
The aforementioned limitations of existing 
inferential methods for MAB collected data create the practical challenge of how to design small pilot studies that can both be highly adaptive and powerful.   
Despite the consensus that pilot studies should focus on feasibility, existing work~\citep{kistin2015pilot} demonstrates that $81\%$ of published pilot experiments incorporated hypothesis testing.
Preliminary positive evidence of an intervention effectiveness increases access to funding; see e.g., the TS study of~\cite{aguilera2020mhealth}, funded by the NIH. 
Conversely, an underpowered study can erroneously discourage promising experimental options progressing to later developmental phases. Reason why, power considerations may prevail over other considerations as a key decision-making element in such exploratory settings. We emphasise that for confirmatory trials type-I error is the most crucial design aspect.

Second, using an exploitative MAB algorithm could help preserve \textbf{participants' engagement} during the course of an e.g., mobile health experiment, and avoid impairing the integrity of collected data~\citep{druce2019maximizing}.
Note that restricting the exploitative nature of the algorithm, as required by current existing inferential approaches, not only leads to substantial regret costs (as shown in Figure~\ref{fig:regretex}) with direct impact on the \textbf{benefits of an experiment's participants}, but it may also indirectly reduce participants' engagement.

\textbf{Our Contribution.} Motivated by the above, in this work, we aim to perform valid and efficient frequentist inference in MAB collected data, without imposing restrictions on the underlying algorithm that may impact its regret optimality. Our contribution is fourfold. First, in Section~\ref{sec:Q-test}, we present a \emph{novel exact statistical test}, that, in virtue of its exactness, is more appropriate than existing asymptotic approaches in small samples. A major advantage of our test is that \emph{it does not require restricting the MAB algorithm} to reduce a potentially aggressive exploitation. Instead, our proposal directly uses the information provided by the adaptive algorithm through the realised \textit{allocation probabilities}, to build the \emph{Allocation Probability Test} (AP-test). As the allocation probabilities used during an experiment are guided by the observed data, they are also informative on the likely values of parameters of interest. More specifically, the more skewed (e.g., towards $1$) are such probabilities, the higher the evidence of superiority of the correspondent arm. Illustratively, the allocation probability observed in our real-world experiment (shown later in Figure~\ref{fig:mturk}) 
suggests a high preference for the experimental arm. 
Such extreme values directly translate into the actual arms' allocations (see also Figure~\ref{fig:regretex}–right plot). However, while offering potential benefit to participants, they also pose challenges for traditional inferential methods, due to a potential high imbalance in arms allocation.  

Second, in Section~\ref{sec:TS-Alloprob}, we \emph{characterize the AP-test for the widely used TS bandit algorithm}, whose optimal (regret) properties have been extensively studied~\citep[see e.g.,][]{chapelle2011empirical,agrawal2013further}. Existing theoretical results on TS allow us to motivate the test for finite samples as well as to derive \emph{asymptotic properties for our test} (more specifically, its statistical power) when applied to TS. 

Third, in Section~\ref{sec:results}, through extensive simulation experiments, we demonstrate the finite-sample inferential performance of our method compared to existing strategies. 
Finally, in Section~\ref{sec:Mturk}, we illustrate its applicability in a real-world field experiment, for understanding users' preferences about messaging options concerning mental health aspects.

\section{Problem Setting}\label{sec:setup}

\paragraph{Setup and Notation.} Suppose there are $T$ interim steps in a study or experiment. At each step $t=0,\dots,T$, participants are accrued in batches of size  $n_t \geq 1$, which can be random or fixed; for ease of exposition, we assume fixed, known sizes (i.e., $n_t = n \: \forall t$), leading to a total experiment or sample size $N = (T+1) \times n$. 
At each step $t$, each participant $i = 1,\dots, n$, can be assigned to only one arm, say $A_{t,i}$, from a discrete set.  We focus on the two-armed setting, i.e., $A_{t,i} \in \{ 0, 1\}$, 
where $0$ and $1$ denote the \textit{control} and the \textit{experimental} arm, respectively. The general ($K+1$) multi-armed setting is illustrated in Supplementary material E. 

Arms are allocated to participants according to a policy $\pi_t \doteq \{\pi_{t,k}, k=0, 1\}$, where each $\pi_{t,k}$ is the allocation probability of arm $k$ to a participant in step $t$. Given the two-armed setting, here we will use as reference the allocation probability of the experimental arm $\pi_{t,1}$, noticing that $\pi_{t,0} = 1-\pi_{t,1}$. These allocation probabilities may be derived from a known (bandit) algorithm targeting a regret-minimization objective–as we will later illustrate for TS–and could in principle depend on covariates (contextual MABs). In a sequential setting, these probabilities are decided at the beginning of each step $t$, only changing from one step to another and remaining constant otherwise.  

According to the potential outcomes framework~\citep{neyman1923} of causal inference, let $Y_{t,i}(A_{t,i})$ be the random \textit{reward} variable representing the outcome that would be observed if participant $i$ were assigned to arm $A_{t,i}$ in step $t$. Denoted by $\mathcal{H}_{t-1} \doteq \{A_{\tau,i}, Y_{\tau,i}(A_{\tau,i}), i=1,\dots, n, \tau = 1,\dots,t-1 \}$ the history of selected arms and associated rewards prior to step $t$, the allocation probabilities are a function of this history, i.e., $\pi_{t,k} \doteq \mathbb{P}(A_{t,i}=k | \mathcal{H}_{t-1})$, for all $k$'s and $t$'s. 

Typically, to guarantee sufficient exploration of arms, restrictions on the maximum ($\pi_{\max}$) and minimum ($\pi_{\min}$) possible values of the allocation probabilities are imposed~\citep[see e.g.,][]{Zhang_NEURIPS2020, yao_power, greenewald2017action}. In such case, $\pi_{t,k} \in [\pi_{\min},\pi_{\max}]$, with $ 0 < \pi_{\min} \leq \pi_{\max} < 1$ and generally $\pi_{\max} = 1- \pi_{\min}$, for all $t$'s and $k$'s. \textbf{In this work, 
we only need the algorithm to be randomized, i.e., $ \boldsymbol{0 < \pi_{t,k} < 1}$ for all $\boldsymbol{k}$'s and $\boldsymbol{t \le T}$}. Note that this less restrictive condition holds for most popular MAB approaches, including TS.

We assume arms' rewards are drawn independently from a fixed distribution with conditional mean:
\begin{align}\label{eq:reward_model}
    \mathbb{E}(Y_{t,i}(A_{t,i})| \mathcal{H}_{t-1}) 
    &=  \mu_{0}(1-A_{t,i}) + \mu_{1}A_{t,i}\\ 
    &= \begin{cases} \mu_0, \quad \text{if}\ A_{t,i} = 0,\\ \mu_1, \quad \text{if}\ A_{t,i} = 1,\end{cases} \nonumber
\end{align}
with $(\mu_0, \mu_1)$ unknown arms' parameters.  
By assuming independent and identically distributed rewards, the conditional mean reward does not depend on the prior history~\citep[\textit{stochastic stationary bandits};][]{lattimore2020bandit}. Here we focus on the stationary case but our approach is largely applicable to non-stationary settings (see Supplementary material D1).

\subsection{Thompson Sampling} \label{sec:TS-Alloprob}

Rooted in a Bayesian framework, the TS algorithm~\citep{thompson1933likelihood} defines an allocation policy by randomizing participants to arms in proportion to the posterior probability of each arm being superior 
at step $t$. In a two-armed setting, denoting by $\pi_{t,1}^{\text{TS}}$ TS's allocation probability of the experimental arm $1$ at step $t$, and considering the reward specifications of Eq.~\eqref{eq:reward_model}, we have that:
\begin{align}\label{eq:TS_alloprob}
     \pi_{t,1}^{\text{TS}} & = \mathbb{P}\Big(\mathbb{E}\big(Y_{t,i}(A_{t,i} = 1)\big) \geq \mathbb{E}\big(Y_{t,i}(A_{t,i} = 0)\big) | \mathcal{H}_{t-1} \Big) \nonumber\\ &= \mathbb{P}\Big(\mu_1 \geq \mu_0 | \mathcal{H}_{t-1} \Big), \quad \quad \quad \quad \forall t \in [0,T]. 
\end{align} 
For some families of reward distributions, it is possible to compute $\pi_{t,1}^{\text{TS}}$ either analytically or by quadrature. However, due to a potential computational burden, 
the typical way of implementing TS~\citep[see e.g., ][]{chapelle2011empirical} avoids the direct computation of such probabilities and proceeds as follows. Assuming a prior distribution for the unknown parameters $(\mu_0, \mu_1)$, at each step $t \ge 1$ a sample is drawn from the updated posterior distributions, say $(\tilde{\mu}_{t,0}, \tilde{\mu}_{t,1})$ and the arm $\tilde{a}_{t,i}$ with the highest estimated posterior mean reward $\mathbb{\tilde{E}}(Y_{t,i}|A_{t,i} = k) = \tilde{\mu}_{t,k}$ is allocated. That is:
\begin{align*}
    \tilde{a}_{t,i} \doteq \text{argmax}_{k = 0, 1}\mathbb{\tilde{E}}(Y_{t,i}|A_{t,i} = k) = \text{argmax}_{k = 0,1}\tilde{\mu}_{t,k}.
\end{align*}
Clearly, for $t=0$ the sample is drawn from the prior. 
If we assume a Gaussian family, with identical priors $N(\mu_\text{prior}, \sigma^2_\text{prior})$, and Gaussian rewards with known and equal variances $\sigma^2_y$ and means specified in Eq.~\eqref{eq:reward_model}, the allocation probability in Eq.~\eqref{eq:TS_alloprob} has closed form:
\begin{align} \label{TS_post}
    \tilde{\pi}_{t,1}^{\text{TS}} = \mathbb{P}(\tilde{\mu}_{t,1} > \tilde{\mu}_{t,0} | \mathcal{H}_{t-1}) = \Phi \left(\frac{\mu_{D_t}}{\sigma_{D_t}} \Bigg| \mathcal{H}_{t-1}\right),
\end{align}
with $\Phi$ indicating the cumulative density function of a Gaussian distribution, and $\mu_{D_t}$ and $\sigma_{D_t}^2$ given by: 
\begin{align} \label{eq:post-param}
    \mu_{D_t} &=\frac{\sigma^2_{y}\mu_{\text{prior}}+\sigma^2_{\text{prior}}\sum_{\tau=0}^{t-1}\sum_{i=1}^{n}Y_{\tau,i}\mathbb{I}(A_{\tau,i}=1)}{\sigma^2_{y} + \sum_{\tau=0}^{t-1}\mathbb{I}(A_{\tau,i}=1)\sigma^2_{\text{prior}}} \nonumber\\
    &- \frac{\sigma^2_{y}\mu_{\text{prior}}+\sigma^2_{\text{prior}}\sum_{\tau=0}^{t-1}\sum_{i=1}^{n}Y_{\tau,i}\mathbb{I}(A_{\tau,i}=0)}{\sigma^2_{y} + \sum_{\tau=0}^{t-1}\mathbb{I}(A_{\tau,i}=0)\sigma^2_{\text{prior}}}; \nonumber\\
    \sigma_{D_t}^2 &= \frac{\sigma^2_{\text{prior}}\sigma^2_{y}}{\sigma^2_{y} + \sum_{\tau=0}^{t-1}\mathbb{I}(A_{\tau,i}=1)\sigma^2_{\text{prior}}} \nonumber\\ &+  \frac{\sigma^2_{\text{prior}}\sigma^2_{y}}{\sigma^2_{y} + \sum_{\tau=0}^{t-1}\mathbb{I}(A_{\tau,i}=0)\sigma^2_{\text{prior}}}.
\end{align}
Proof is given in Supplementary material A1.

When $\mu_1>\mu_0$ the distribution of $\pi_{t,1}^{\text{TS}}$ is skewed towards $1$, and converges to $1$ as $t \to \infty$~\citep{kalkanli2020asymptotic}. This property explains the large imbalance between arms' sample sizes, generally favouring arm $1$. 
When $\mu_1=\mu_0$, the distribution of $\pi_{t,1}^{\text{TS}}$ does not concentrate~\citep[see e.g.,][]{Zhang_NEURIPS2020} and may result (though less likely) in extremely imbalanced allocations as well, favouring either arm $1$ or $0$. Supplementary material A4 provides illustrative examples. 
Such highly exploitative behaviour of TS negatively impacts the theoretical asymptotic distribution of traditional statistical estimators, motivating the need for specific TS-collected data techniques. 

\subsection{Hypothesis Testing for Superiority} \label{sec:hypo-testing}
Using data collected through a MAB-driven experiment, like TS, 
our inferential problem is to decide whether a new experimental arm is superior to a control one. More formally, we want to test a \textit{null} $H_0$ versus an \textit{alternative hypothesis} $H_1$, specifically defined  as $H_0\!: \mu_1 = \mu_0$ vs. $H_1\!: \mu_1 > \mu_0$.

A standard statistical approach for hypothesis testing uses an appropriate \textit{test statistic} $T(Y)$ which is a function of the reward data $Y$~\citep{casella2021statistical}. 
For example, the approaches of \cite{Zhang_NEURIPS2020} and \cite{hadad2021confidence} are based on a modified version of the Z-test (traditionally measuring the distance from two sample means in relation to their standard errors), in which the novel BOLS and AW-AIPW estimators, respectively, replace the sample means.

Such a statistic is typically used to form a decision rule that targets a given probability $\alpha$ of rejecting the null $H_0$ when it is actually true (i.e., \textit{type-I error} probability), by selecting a \textit{critical value} $t_\alpha^*$. More specifically, 
\begin{align*}
    t_\alpha^*\!:\ \mathbb{P}\left(\text{Reject}\ H_0 | H_0\right) = \mathbb{P}\left(T(Y) > t_\alpha^* | \mu_1 = \mu_0 \right) = \alpha.
\end{align*}
Given $t_\alpha^*$, 
$T(Y)$ should also ensure a low risk of failing to reject $H_0$ under the alternative (i.e., low \textit{type-II error} $\beta$), or alternatively a high
\textit{power}, i.e., the probability of correctly rejecting the null, defined as:
\begin{align*}
    \mathbb{P}\left(\text{Reject}\ H_0 | H_1\right) = \mathbb{P}\left(T(Y) > t_\alpha^* | \mu_1 > \mu_0 \right) =1-\beta.
\end{align*}
Typically, one would like to design an experiment with high power, while controlling type-I error at a level $\alpha$, e.g., $0.05$. However, for a fixed experiment of size $N$, low values of $\alpha$ translate into reduced power. 
This explains why small-sample settings may target a higher $\alpha$ as a way to achieve higher power~\citep{schoenfeld1980statistical}. 

\section{The Allocation Probability Test}\label{sec:Q-test}

A distinctive feature of our \textit{Allocation Probability Test}, or AP-test, 
is that it is defined as a function of the allocation probabilities only. 
It thus requires only minimal knowledge, namely, the adaptive algorithm used to determine the allocation probabilities for collecting the data, and the observed arms' allocation probabilities during the experiment, but not the reward data. 

Our work is inspired by a recent idea proposed in~\cite{barnett2020novel}, where a similar test is implemented for the \textit{covariate-adjusted Forward Looking Gittins Index}~\citep[FLGI;][]{villar2018covariate}, in a two-armed binary reward setting. In our work, we generalize the allocation probability-based test to be used with any generic randomized bandit algorithm, and we then focus on its theoretical and empirical properties when applied to TS. Notably, in addition to being a widely used and highly interpretable algorithm, we show how TS's optimal theoretical properties~\citep{chapelle2011empirical, agrawal2012analysis, agrawal2013further} naturally translate into theoretical properties for our proposed AP-test. Such theoretical properties are not available for the FLGI  in~\cite{barnett2020novel}.


\subsection{General Allocation Probability Test} 
In a general two-armed adaptive setting, we define the Allocation Probability Test statistic AP$_T$ as:
\begin{align}\label{eq:Pi-test}
    \text{AP}_T \doteq 
    \sum_{t=t_{\min}}^T\mathbb{I}\left(\pi_{t,1} > 0.5 \right), 
\end{align}
where $t_{\min}$ accounts for a minimum data requirement to inform $\pi_{t,1}$ updates. We set $t_{\min} = 1$, discarding $\pi_{0,1}$ which is not informed by the current study's data.

Conceptually, the test is defined as the number of steps $t \leq T$, for which the allocation probability of the experimental arm $\pi_{t,1}$ exceeds that of a (non-adaptive) equally randomized (ER) experiment, i.e., $\pi^{\text{ER}} = 0.5$. 
For a general $(K+1)$-armed case, $\pi^{\text{ER}} \doteq 1/(K+1)$, as discussed in the Supplementary material F3. 

The rationale for the test comes from the fact that the allocation probabilities of an \emph{aggressively} adaptive algorithm are designed to quickly 
favour the most promising arm. Thus, these probabilities will differ the most from those in an ER experiment, the further we are from $H_0$. 
Thus, the higher the evidence of superiority of arm 1, the  higher $\pi_{t,1}$ at each $t$, and the higher the value of the AP-test.

In order to use this test for hypothesis testing, similarly to the standard testing approach in Section~\ref{sec:hypo-testing}, we need to derive its distribution under the null, and an associated critical value, 
defined as:
\begin{align}\label{eq:Pi_quantile}
    q_\alpha^* \doteq \min \{q\in [0,\dots,T]\!: \mathbb{P}(\text{AP}_T > q | H_0) \leq \alpha \},
\end{align}
Given the discrete nature of the AP-test and its distribution (Figure~\ref{fig:Q_distribution} provides an illustration), deriving the critical value using fixed standard values of $\alpha$ such as $0.05$, may result in highly conservative tests, i.e., with a rejection rate well below the nominal level $0.05$~\citep{yates1984tests,little1989testing}, including $0$. To avoid a $0$ type-I error, that will translate in a $0$ power (when, according to Eq.~\eqref{eq:Pi_quantile}, the resulted $q_\alpha^* = T$), in such case only, we implement the AP-test with the critical value $q_\alpha^* = T-1$. We quantify the type-I error accordingly, noticing that $T-1$ represents the critical value that will result in the minimum strictly positive error.

For ensuring proper type-I error control, there are two alternatives: (1) to implement the test with non-standard but pre-determined $\alpha$ levels~\citep[see e.g.,][]{schoenfeld1980statistical}, or (2) to ensure type-I error control at any $\alpha$ level by applying a randomized-based procedure (see Supplementary material B). In this paper, we will present results of both of these approaches.  


\subsection{TS Allocation Probability Test}
The AP-test can be in principle applied to any randomized algorithm, and its distribution directly depends on the allocation probabilities determined by the underlying algorithm, according to the observed reward data. 
Here we illustrate properties of the allocation test when applied to TS, whose asymptotic properties are used to derive properties of the theoretical distribution of the AP-test. 
Notably, for algorithms whose asymptotic convergence to the optimal arm can be ensured (i.e., $\pi_{t,1}\to 1$ as $t\to \infty$), as in the case of TS,  
our proposed test finds an ideal fit.  
Notice that in the two-armed case, $H_0$ can be thought of as two non unique optimal arms and the case of $H_1$ as a unique optimal arm, and the question is to differentiate between these two cases with our test.

Because of the discrete nature of the AP-test, its exact distribution  can be derived by simply computing the probability that the test takes values $0,\dots, T$.
In practice, for doing inference it typically suffices to compute 
$\mathbb{P}(\text{AP}_T > T-1) = \mathbb{P}(\text{AP}_T = T)$. 
Under TS and the Gaussian setting assumed in Section~\ref{sec:TS-Alloprob}, we compute this probability mass in closed-form, resulting in:
\begin{align} \label{eq:exact_typeI}
    \mathbb{P}(\text{AP}^{\text{TS}}_T = T) &= \mathbb{P}\left(\sum_{t=1}^T\mathbb{I}(\pi_{t,1}>0.5)=T\right)\\ 
    & = \frac{1}{2}\left[ \Phi \left( \frac{\mu_1}{\sigma_y}\right)  + 1 - \Phi \left( \frac{\mu_0}{\sigma_y}\right)\right] \times \nonumber\\
    \times \prod_{t=2}^T &\mathbb{P}\left( \frac{\mu_{D_t}}{\sigma_{D_t}}>0 \Big| \frac{\mu_{D_{t-1}}}{\sigma_{D_{t-1}}}>0,\dots,\frac{\mu_{D_1}}{\sigma_{D_1}}>0\right), \nonumber
\end{align}
with $\mu_{D_t}$ and $\sigma_{D_t}$ as in Eq.~\eqref{eq:post-param}. 
Proof, given in Supplementary material A2, also shows how, by applying the law of total probability, conditioning on all possible arms sequences, each $\mu_{D_t}/\sigma_{D_t}$ is simply a sum of Gaussian variables, allowing for an exact solution.

\begin{figure}[ht]
    \centering
    \includegraphics[scale = .35]{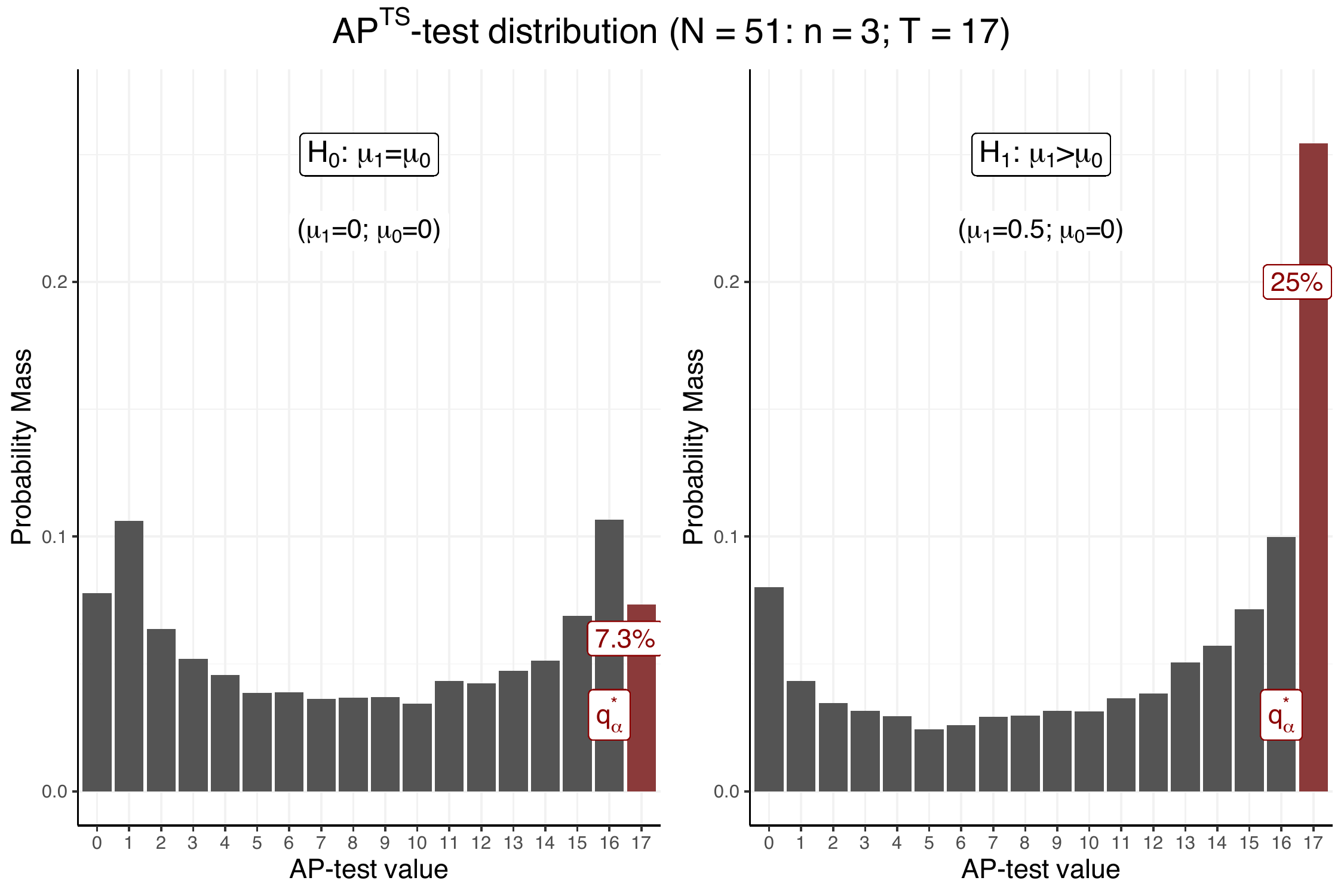}
    \caption{TS AP-test approximate distribution\protect\footnotemark~in a two-armed setting with rewards: i) $Y_{t,i} (A_{t,i}=k) \sim N(0, \sigma^2 = 10)$ ($H_0$; left plot), $\forall k$ and, ii) $Y_{t,i} (A_{t,i} = 0) \sim N(0, \sigma^2 = 10)$ and $Y_{t,i} (A_{t,i} = 1) \sim N(0.5, \sigma^2 = 10)$ ($H_1$; right plot), $\forall t,i$. Type-I error and power (left and right red bar, respectively) are based on $\alpha = 0.1$.
    }
    \label{fig:Q_distribution}
\end{figure}

\textbf{Remark.} The probability in Eq.~\eqref{eq:exact_typeI} is of key importance as for $T < 1000$, we found that $T-1$  typically is the test's critical value $q_\alpha^*$, and, thus, Eq.~\eqref{eq:exact_typeI} represents either the exact type-I error (under $H_0$) or exact power (under $H_1$) of the AP-test (see also Figure~\ref{fig:Q_distribution}). 

\setcounter{footnote}{-1}

\begin{figure}[ht]
    \centering
    \includegraphics[scale = .47]{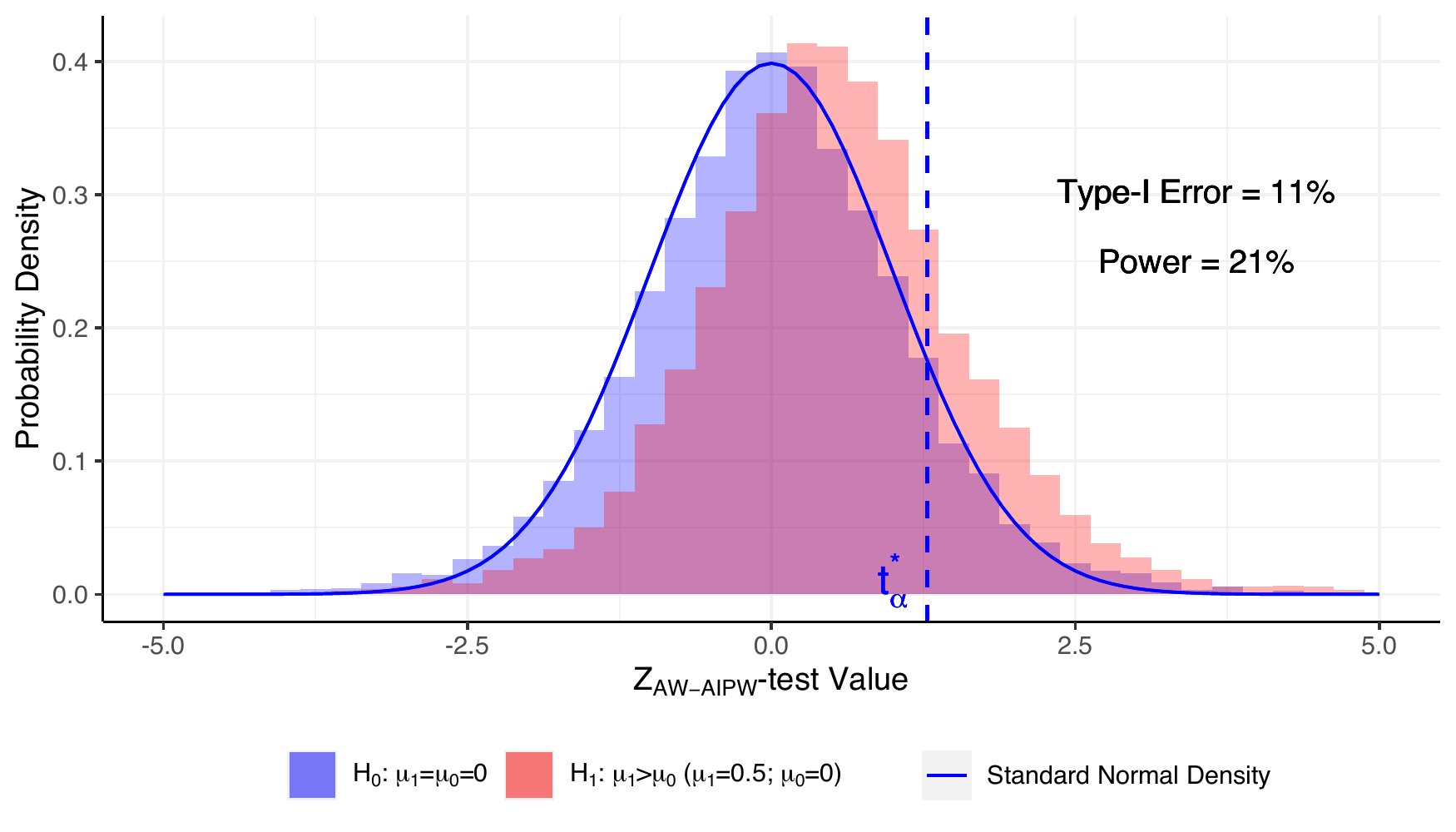}
    \caption{Theoretical (standard Gaussian) and approximate and distribution of the AW-AIPW-based Z-test in a two-armed setting with data generated and inference performed according to details of Figure~\ref{fig:Q_distribution}.}
    \label{fig:distr-awaipw}
\end{figure}

As shown in Figure~\ref{fig:Q_distribution}, the Thompson sampling AP-test distribution under $H_0$ (top plot) is a symmetric–around the midpoint of $(T/2 + 1)$–bi-modal discrete distribution with support $\{0,1,\dots, T\}$; while, under $H_1$ (bottom plot), it is a left-skewed distribution. 
\footnotetext{Exact computation of the AP-test distribution (according to Eq.~\eqref{eq:exact_typeI}, can be very intensive (e.g., for large $T$ and $N$), but is nonetheless possible. In line with common practice in finite samples~\citep[see e.g.,][]{ericsson1991monte}, we adopt a Monte Carlo (MC) procedure for approximating it.}
Differently from traditional asymptotic test statistics, such as the Z-tests, the distribution of the AP-test changes sharply with $T$, and so does the critical value $q_\alpha^*$. In the illustrative example with $N=51$ and $T=17$, the AP-test assumes values $\{0,1,\dots, T=17\}$, and, if we define $q_\alpha^*$ according to $\alpha = 10\%$ (in line with a motivating small-samples pilot study), this is given by $T-1 = 16$. The red bars corresponding to $T=17$ represent the associated type-I error ($\mathbb{P}(\text{AP}^{\text{TS}}_T > q_\alpha^* | H_0) = 0.073$; left plot) and power ($\mathbb{P}(\text{AP}^{\text{TS}}_T > q_\alpha^* | H_1) = 0.25$; right plot). This power level illustrates the efficiency gain of the AP-test, as existing approaches in the same setting ($N=51$) have lower power even with type-I error above nominal level (see Figure~\ref{fig:distr-awaipw} for a visual comparison of traditional tests and our proposed approach). 

\textbf{Remark.} TS's asymptotic behaviour and its different convergence rates for the case of multiple optimal arms ($\mu_0 =\mu_1$) vs one superior arm ($\mu_0 <\mu_1$ or $\mu_0 >\mu_1$), ensures that, for a given $T$,  the mass in the discrete distribution of the AP-test for the value $T$ will be systematically higher in the latter scenario (power) than the former (type-I error). Using the asymptotic convergence optimality results of TS~\citep{kalkanli2020asymptotic}, allows us to formulate the following theorem. 
\begin{theorem}
If $\mu_1 > \mu_0$, and arm's allocation probabilities are defined according to standard Thompson Sampling as in Eq.~\eqref{TS_post}, then the allocation probability $\text{AP}$-test diverges as $T \to \infty$, i.e., $\text{AP}^{\text{TS}}_T \to \infty$.
\end{theorem}

Proof is based on the divergence test and \textit{Lemma 3.2} on TS, both reported in the Supplementary material A3. Note that this theorem easily extends to a general $(K+1)$-arm TS setting with a unique optimal arm, and to other adaptive algorithms with the allocation probability of the optimal arm converging to $1$. 

\begin{lemma}
If $\mu_1 > \mu_0$, 
TS converges to allocating the arm $1$ with probability $1$ as $t \to \infty$, i.e.,
\begin{align*}
    \lim_{t \to \infty} \pi_{t,1}^{TS} = 1,\quad \text{when}\  \mu_1 > \mu_0.
\end{align*}
\end{lemma}

These theoretical results allow us to outline the following properties of the $\text{AP}^{\text{TS}}_T$-test, also supported by empirical evaluations in Supplementary material A4. 

First, \textbf{the sample size imbalance between arms caused by an aggressive nature of adaptive algorithms 
does not negatively affect the statistical power of the AP-test}. 
To the contrary, the larger the mean difference ($\mu_1-\mu_0$), the larger the expected sample size imbalance and the 
more the allocation probabilities are likely to be greater than the threshold $\pi^{\text{ER}} = 0.5$. 
Thus,  the larger $\mu_1-\mu_0$ is, the more skewed the $\text{AP}$-test distribution under $H_1$ (as also shown in Figure~\ref{fig:Q_distribution}; right plot). 

Second, since power is not decreased by a potentially highly exploitative nature of the adaptive algorithm, \textbf{the power of this test does not improve by constraining the arms' allocation probabilities}, as in the approaches of~\cite{Zhang_NEURIPS2020} and \cite{hadad2021confidence}; see also results in Supplementary material A4. 

Finally, while in principle $\text{AP}^{\text{TS}}_T$ could diverge much slower than $T$ and $q^*_\alpha$ scales with $T$, the different convergence rate of TS and its optimality under the alternative, ensures that our test will discriminate (increasingly better) between the alternative and the null. This makes the $\text{AP}$-test a powerful approach, not only in finite samples, but also asymptotically. Note that, while $\text{AP}^{\text{TS}}_T$ \textbf{diverges under the alternative} (with a rate depending on the actual arms differences, resembling TS's theoretical properties), \textbf{the divergence is not verified under the null}.

\section{Simulation Studies}\label{sec:results}

\paragraph{Experiments.} 
For our simulation studies, we focus on the setting introduced in Section~\ref{sec:setup}, and consider the following different scenarios:
\begin{enumerate}[noitemsep,topsep=0pt]
    \item[(i)] \textit{fully-sequential experiments}: maximum sample size $N=150$, $t \in [0,150]$, and batch size $n=1$;
    \item[(ii)] \textit{small-batched experiments}: maximum sample size $N=150$, $t \in [0,50]$, and batch size $n=3$;
    \item[(iii)] \textit{medium-batched experiments}: maximum sample size $N=150$, $t \in [0,15]$, and batch size $n=10$.
\end{enumerate}
We are interested in the case where, for cost or other reasons, we are limited by a maximum sample size $N$. We set $N=150$, driven by our real-world experiment, where each participant received a monetary budget for their participation. Then we let the batch size and number of batches vary accordingly. 

We assess the performances of our test compared to state-of-the-art strategies valid under TS policies (i.e., BOLS and AW-AIPW; details reported in the Supplementary material C). We consider the following hypotheses: $H_0\!: \mu_0 = \mu_1 = 0$ and $H_1\!: \mu_1 > \mu_0$, with $\mu_1 = 0.5$ and $\mu_0 = 0$, and set $\alpha = 0.05$. We generate data according to TS, assuming Gaussian reward distributions with equal variances ($\sigma^2 = 10$), and means specified by the aforementioned hypotheses. For both arms, we choose $N(0, 10)$ priors. Note that, in order to fulfil the underlying assumptions of the comparative methods, we use restricted TS (i.e., with constraints on the allocation probabilities–as discussed in the original works and illustrated in Supplementary material C3); for the AP-test, we use standard TS.
\paragraph{Type-I Error Control and Adjustment.} 
\begin{figure}[ht]
    \centering
    \includegraphics[scale = .47]{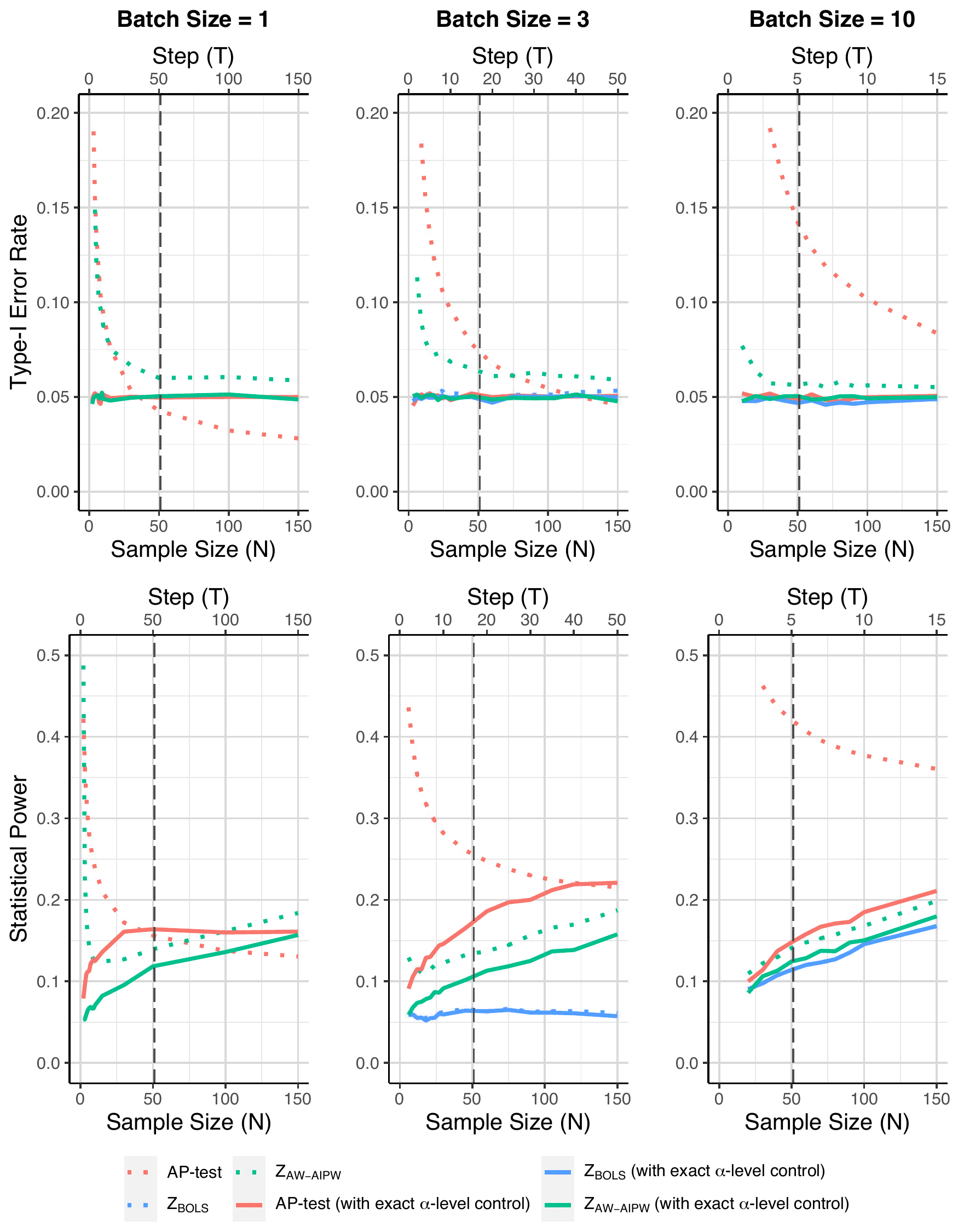}
    \caption{Type-I error and power of inferential approaches in different batch sizes. 
    We also report type-I error and 
    power when exact $\alpha = 0.05$ is targeted. Values are obtained by averaging across $10^4$ independent TS trajectories, with standard errors always $< 0.005$.}
    \label{fig:typeI_power}
\end{figure}

As discussed in Sections~\ref{sec:setup} and \ref{sec:Q-test}, in general we want to control type-I error at some level $\alpha$ and then maximize power. While the exact type-I error control difficulty typical of discrete tests was discussed in Section~\ref{sec:Q-test}, we note that, despite the theoretical guarantees, existing asymptotic approaches may also suffer from such issue, particularly in small samples (see e.g., the AW-AIPW in Figure~\ref{fig:typeI_power}). To ensure a higher comparability in terms of power, we adopt a randomized procedure that is able to perfectly control type-I error (as shown in Figure~\ref{fig:typeI_power}). We provide such procedure for discrete tests in Supplementary material B, and extend it to continuous tests, which are not exempt from the aforementioned problem. Note that, despite being only an auxiliary result, we include this extension among the contributions of this work. Such procedure ensures the applicability of any test in situations where exact control may be needed. We also emphasize that for $t \geq 40$, our discrete test is able to largely control type-I error at the usual $0.05$-level, as it is a decreasing function of $t$ (as demonstrated in Supplementary material A4). 

\paragraph{Statistical Power.} Comparative power results, adjusted for exact type-I error control as well (with $\alpha$ = 0.05), are reported in Figure~\ref{fig:typeI_power}. In addition to the overall regret trade off (as illustrated in Figure~\ref{fig:regretex}), we see that the BOLS and AW-AIPW asymptotic approaches exhibit a very low power in small samples, which makes them unsuitable for pilot experiments. This is particularly true for the BOLS method in the fully-sequential ($n=1$) or small-batched setting ($n=3$), where it is either inapplicable or with no power efficiency, due to its theoretical guarantees for $n \to \infty$. On the contrary, the AW-AIPW approach does not seem to be highly affected by batch size but sample size only, while our approach shows a dependency on both elements. In the evaluated scenarios, AP-test's performances are always higher compared to the other methods; for higher batch sizes, power performances are comparable (see Supplementary material D). Its advantages are remarkable in small batch sizes ($n=3$), where, based on a sample size of $N=51$ for example, it achieves a power of $25\%$, compared to the $14\%$ and $6\%$ of AW-AIPW and BOLS, respectively. Note that this is the most common setting in applications that come with costs in terms of sample sizes, and where we also see the highest value of our procedure. In large batch sizes–less common in practical applications–all methods have comparable power (see Supplementary material D2). Notice, however, that our method comes always with an improved regret performance.

\section{Real-world MTurk Experiment} \label{sec:Mturk}
\begin{figure}[ht]
    \centering
    \includegraphics[scale = 0.44]{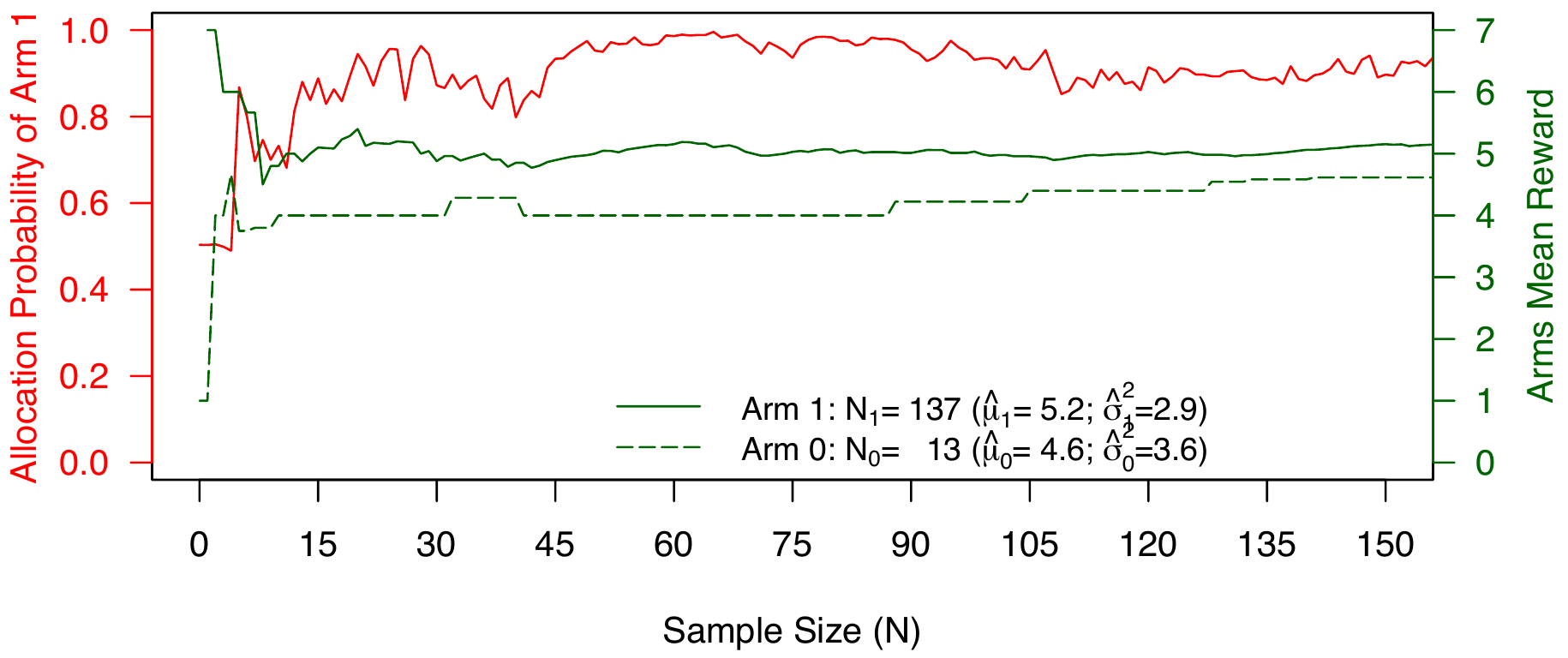}
    \caption{Summary of the two-armed MTurk experiment, with data collected through the TS algorithm.}
    \label{fig:mturk}
\end{figure}

The challenges of performing traditional inference on data collected by regret-minimizing MAB algorithms can become a real barrier for either applying these algorithms in many real-world settings or using those data for making inference. To illustrate this, we conducted a TS experiment using the Amazon Mechanical Turk~\citep[MTurk; see e.g.,][]{mason2012conducting}, a popular \textit{crowdsourcing} site among social scientists for running experiments. Participants were asked to provide a rating from 0 to 7 for two types of messages in terms of how helpful these would be in terms of helping them manage their mood. A total of $N=150$ participants were recruited online in batches of prespecified size $n=3$. A summary of the reward data (ratings of each arm) and the observed allocation probabilities trajectories are reported in Figure~\ref{fig:mturk}. 
The evidence-suggested superiority of arm 1 (
$\hat{\mu}_1 = 5.2$ vs $\hat{\mu}_0 = 4.6$), quickly skewed the TS allocation probability towards 1, leading to a total of $N_1 = 137$ vs $N_0 = 13$ participants allocated to arm 1 and arm 0, respectively. Estimated mean rewards are consistent with an independent ER experiment run in MTurk. As the observed allocation probabilities of the TS are always $>0.5$, our proposed AP-test rejects the null with any $\alpha > 0.05$ (as shown in Figure~\ref{fig:typeI_power}, type I error is always controlled at $\alpha = 0.05$ when $N \geq 150$). In this case, the extreme imbalance between the two arms, would have made the analysis of these data with traditional approaches completely underpowered, in addition to invalid (due to the use of the standard, unrestricted, TS). 

\begin{figure}[ht]
    \centering
    \includegraphics[scale = 0.7]{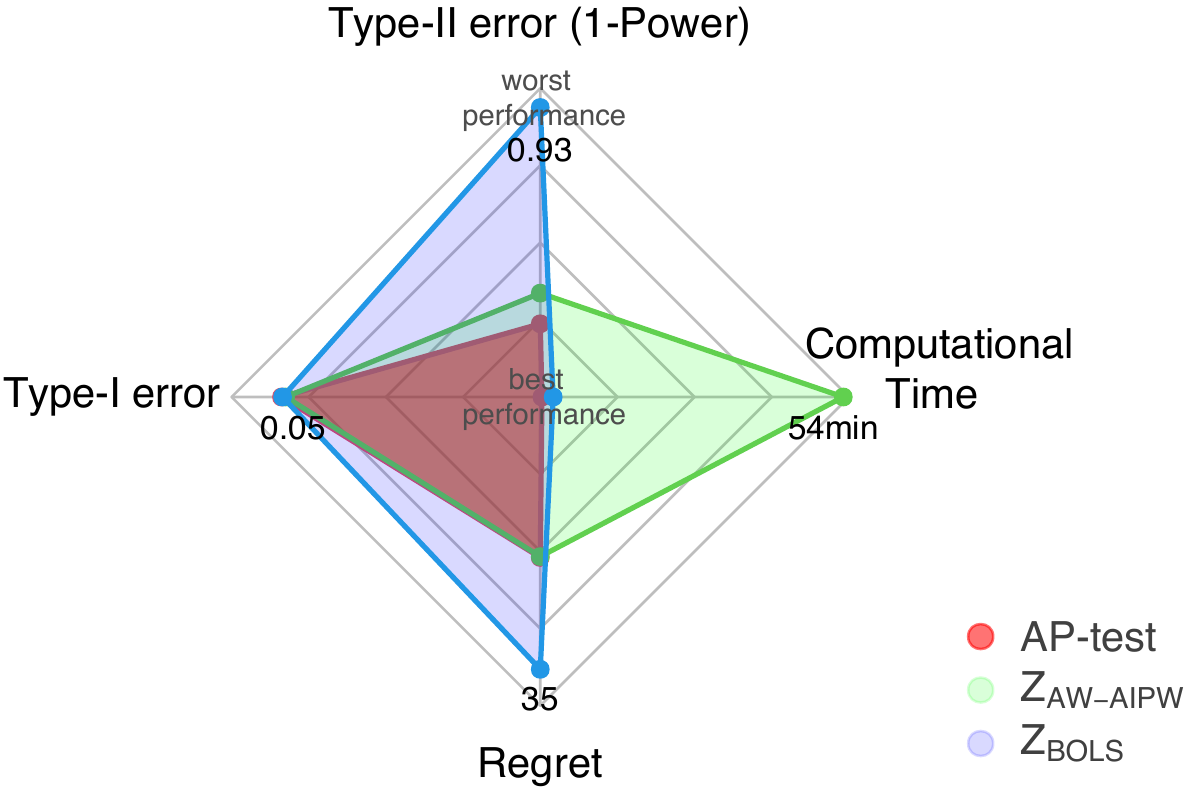}
    \caption{Radar chart comparing performances of the three methods according to the real-world study setting ($N=150$, $n=3$, $\hat{\mu}_1 = 5.2$, $\hat{\mu}_0 = 4.6$ and $\sigma^2 = 4$). Smaller areas correspond to better performance.}
    \label{fig:radarplot}
\end{figure}

In Figure~\ref{fig:radarplot} we propose a global comparison among the different strategies considering our experimental setting. In addition to regret and inferential properties, we include the computational time for doing inference, specifically for computing type-I error (or power) across $10^4$ TS trajectories. Note that, while the computational times are not relevant for hypothesis testing on a specific dataset, for design considerations, especially when aiming for exact type-I error control, an entire set of MC samples are required. Compared to the AP-test and the BOLS, AW-AIPW has the worst performance, as its computation involves estimating an augmented part for the reward model, in addition to certain adaptive weights, for each new observation (see Supplementary material C2), increasing thus the computational complexity of the method.

For reproducibility of the experiments, code and data will be made available (Supplementary material F).

\section{Discussion and Conclusion}

We introduce a general class of exact statistical tests based on the allocation probabilities of the underlying adaptive algorithm, and discuss its applicability and properties in relation to TS. Our proposal represents an innovative way to perform hypothesis testing and an alternative to traditional approaches. The novelty comes from building a test statistic by   making direct use of  
the algorithm that was used to collect the data. 
Compared to state-of-the-art approaches, our proposal: i) does not impose limits on the exploitative nature of the algorithm, ii) can be applied in fully-sequential settings (i.e., with allocation probabilities updated after each observation), iii) especially important, can be a more efficient alternative for small (pilot) studies, given its increased power performance. In summary, our proposed exact approach preserves the regret optimalities of MAB algorithms, while improving statistical power in relation to existing approaches, in different small sample scenarios (as shown with extensive simulations). 
Additionally, with a real-world study, we illustrate its applicability in a setting where alternative approaches are underpowered, in addition to being not applicable. We finally emphasize the extreme flexibility of our approach, suitable in both stationary and non-stationary settings, and that would be easily implementable in more complex settings, e.g., with non-fixed batch sizes or contextual information~\citep[as in][]{barnett2020novel}. Open questions and further work is required to understand how the extension to those contexts would perform in practice. 

\paragraph{Acknowledgments}
This research was supported by the NIHR Cambridge Biomedical Research Centre (BRC-1215-20014) for Nina Deliu, and by the UK Medical Research Council (MC UU 00002/15) for Sofia S. Villar. The views expressed are those of the authors and not necessarily those of the NIHR or the Department of Health and Social Care. Authors thank Audrey Durand for the constructive feedback, and Taneea S. Agrawaal, Harsh Kumar, Jiakai Shi and Tong Li for their help in the MTurk deployments.

\bibliography{main_paper.bib}

%

%

\onecolumn
\aistatstitle{Supplementary Material to Manuscript:\\ Efficient Inference Without Trading-off Regret in Bandits:\\ An Allocation Probability Test for Thompson Sampling
}

The outline of this Supplementary material is structured as follows:
\begin{outline}[enumerate]
   \1 Various Results and Properties of TS-based Allocation Probability Test
    \2 Exact computation of TS's allocation probabilities
    \2 Exact distribution of the TS-based AP-test
    \2 Proof of Theorem 3.1 and Lemma 3.2
    \2 Characteristics and Properties of TS's Allocation Probabilities and TS-based AP-test
   \1 Randomization Procedure for Exact Type-I Error Control
    \2 Discrete Tests
    \2 Continuous Tests
   \1 Comparative Testing Approaches
    \2 Batched Ordinary Least Square (BOLS)-based Z-test
    \2 Adaptively Weighted Augmented Inverse Probability Weighting (AW-AIPW)-based Z-test
    \2 Restricted (BOLS and AW-AIPW) Thompson Sampling
   \1 Additional Experiments
    \2 Sensitivity to non-stationarity
    \2 Type-I Error and Power in higher batch sizes
   \1 Extensions to a General $(K+1)$ Multi-Armed Bandit Setting
    \2 General Setup
    \2 Multiple Hypothesis Testing Problem
    \2 Thompson Sampling Allocation Probability Test
   \1 Code and Data for Results Reproducibility
   
\end{outline}

\newpage

\section*{A. Various Results and Properties of TS-based Allocation Probability Test}

In this section we provide proofs and derive properties of the allocation probability test when applied to Thompson Sampling (TS) focusing on the two-arm case. Illustrative examples support the theoretical findings.


\subsection*{A1. Exact computation of TS's allocation probabilities}\label{sec: exact_TS}

The analytical computation of the allocation probabilities of Thompson Sampling is possible for some families of reward distributions. We now illustrate it in a two-arm setting (arms $A_{t,i} \in \{0,1\}$ for all $t=0,\dots, T$ and $i = 1,\dots,n$) with Gaussian rewards. 

Assuming the following distributions for the arms' rewards:
\begin{align} \label{eq:gaussian-setting}
    Y_{t,i} (A_{t,i} = 0) &\sim \mathcal{N}(\mu_0, \sigma^2_y), \quad \forall t, i, \nonumber\\
    Y_{t,i} (A_{t,i} = 1) &\sim \mathcal{N}(\mu_1, \sigma^2_y), \quad \forall t, i, \nonumber\\
    \mu_0, \mu_1 &\sim \mathcal{N}(\mu_{\text{prior}}, \sigma^2_{\text{prior}}),
\end{align}
with $(\mu_0, \mu_1)$ the unknown arm means parameters, we have that TS's allocation probabilities~\citep[according to iterative procedure in its original formulation; see e.g., ][]{thompson1933likelihood, agrawal2012analysis} at each step $t$ and for each arm are given by:
\begin{align*}
     \tilde{\pi}_{t,1}^{\text{TS}} &= \mathbb{P}(\tilde{\mu}_{t,1} > \tilde{\mu}_{t,0} | \mathcal{H}_{t-1}) = \mathbb{P}(\tilde{\mu}_{t,1} - \tilde{\mu}_{t,0} > 0 | \mathcal{H}_{t-1}) = \Phi \left(\frac{\mu_{D_t}}{\sigma_{D_t}} \Bigg| \mathcal{H}_{t-1}\right)\\
     \tilde{\pi}_{t,0}^{\text{TS}} &= \mathbb{P}(\tilde{\mu}_{t,0} > \tilde{\mu}_{t,1} | \mathcal{H}_{t-1}) = 1 - \tilde{\pi}_{t,1}^{\text{TS}} = 1 - \Phi \left(\frac{\mu_{D_t}}{\sigma_{D_t}} \Bigg| \mathcal{H}_{t-1}\right) = \Phi \left(-\frac{\mu_{D_t}}{\sigma_{D_t}} \Bigg| \mathcal{H}_{t-1}\right),
\end{align*}
with $\mu_{D_t}$ and $\sigma_{D_t}$ the mean and variance of the difference in the updated posterior distributions at each step $t$, as we demonstrate below in Eq.~\eqref{eq:TS_mu} and Eq.~\eqref{eq:TS_sigma}. 

\paragraph{Proof} Note that, at each step $t$, the posterior means $\tilde{\mu}_{t,1}$ and $\tilde{\mu}_{t,0}$ are independent random variables with the following normal distributions:
\begin{align*}
    \tilde{\mu}_{t,1} | \mathcal{H}_{t-1} &= \mathcal{N}\left(\frac{1}{\frac{1}{\sigma^2_{\text{prior}}}+\frac{N_1}{\sigma^2_{y}}}\left(\frac{\mu_{\text{prior}}}{\sigma^2_{\text{prior}}}+\frac{\sum_{\tau=1}^{t-1}\sum_{i=1}^{n}Y_{\tau,i}\mathbb{I}(A_{\tau,i}=1)}{\sigma^2_{y}}\right), \left(\frac{1}{\sigma^2_{\text{prior}}}+\frac{N_1}{\sigma^2_{y}}\right)^{-1} \right),\\
    \tilde{\mu}_{t,0} | \mathcal{H}_{t-1} &= \mathcal{N}\left(\frac{1}{\frac{1}{\sigma^2_{\text{prior}}}+\frac{N_0}{\sigma^2_{y}}}\left(\frac{\mu_{\text{prior}}}{\sigma^2_{\text{prior}}}+\frac{\sum_{\tau=1}^{t-1}\sum_{i=1}^{n}Y_{\tau,i}\mathbb{I}(A_{\tau,i}=0)}{\sigma^2_{y}}\right), \left(\frac{1}{\sigma^2_{\text{prior}}}+\frac{N_0}{\sigma^2_{y}}\right)^{-1} \right),
\end{align*}
with $N_1$ and $N_0$ the total number of times arm $0$ and arm $1$ were allocated up to step $t$, respectively.
The derivation follows from the conjugate normal family, that allows analytically tractable posteriors, with: i) mean given by a weighted mean of the prior and the collected evidence (likelihood) means, and weights dictated by their precisions (reciprocal of their variances), and ii) variances given by the reciprocal of the sum of their precisions. 

Now, since we are interested in the posterior parameters' difference, based on the properties of sums/differences of independent normal variables, we have that 
$\tilde{\mu}_{t,1} - \tilde{\mu}_{t,0} | \mathcal{H}_{t-1} = \mathcal{N}\left( \mu_{D_t}, \sigma_{D_t}^{2}\right)$, where $\mu_{D_t}$ and $ \sigma_{D_t}^{2}$ are given by:
\begin{align}
    \mu_{D_t} =
    &\frac{1}{\frac{1}{\sigma^2_{\text{prior}}}+\frac{N_1}{\sigma^2_{y}}}\left(\frac{\mu_{\text{prior}}}{\sigma^2_{\text{prior}}}+\frac{\sum_{\tau=1}^{t-1}\sum_{i=1}^{n}Y_{\tau,i}\mathbb{I}(A_{\tau,i}=1)}{\sigma^2_{y}}\right) \label{eq:TS_mu} \\ &- \frac{1}{\frac{1}{\sigma^2_{\text{prior}}}+\frac{N_0}{\sigma^2_{y}}}\left(\frac{\mu_{\text{prior}}}{\sigma^2_{\text{prior}}}+\frac{\sum_{\tau=1}^{t-1}\sum_{i=1}^{n}Y_{\tau,i}\mathbb{I}(A_{\tau,i}=0)}{\sigma^2_{y}}\right) \nonumber\\
    = &\frac{\sigma^2_{y}\mu_{\text{prior}}+\sigma^2_{\text{prior}}\sum_{\tau=1}^{t-1}\sum_{i=1}^{n}Y_{\tau,i}\mathbb{I}(A_{\tau,i}=1)}{\sigma^2_{y} + N_1\sigma^2_{\text{prior}}} \nonumber\\
    &- \frac{\sigma^2_{y}\mu_{\text{prior}}+\sigma^2_{\text{prior}}\sum_{\tau=1}^{t-1}\sum_{i=1}^{n}Y_{\tau,i}\mathbb{I}(A_{\tau,i}=0)}{\sigma^2_{y} + N_0\sigma^2_{\text{prior}}} \nonumber\\
    & \nonumber\\
    \sigma^2_{D_t} =
    & \left(\frac{1}{\sigma^2_{\text{prior}}}+\frac{N_1}{\sigma^2_{y}}\right)^{-1} +  \left(\frac{1}{\sigma^2_{\text{prior}}}+\frac{N_0}{\sigma^2_{y}}\right)^{-1} \label{eq:TS_sigma}\\
    = &\frac{\sigma^2_{\text{prior}}\sigma^2_{y}}{\sigma^2_{y} + N_1\sigma^2_{\text{prior}}} +  \frac{\sigma^2_{\text{prior}}\sigma^2_{y}}{\sigma^2_{y} + N_0\sigma^2_{\text{prior}}}. \nonumber
\end{align}
Note that for $t=0$, the two quantities are only based on the prior specification.

Finally, by virtue of the symmetry of Gaussian distributions, we have that:
\begin{align}\label{eq: exact_TS}
    \tilde{\pi}_{t,1}^{\text{TS}}
    &= \mathbb{P}\left(\tilde{\mu}_{t,1} - \tilde{\mu}_{t,0} > 0 | \mathcal{H}_{t-1}\right)
    = \mathbb{P}\left(\mathcal{N}\left( \mu_{D_t}, \sigma_{D_t}{2}\right)>0| \mathcal{H}_{t-1}\right) \\ 
    &= \mathbb{P}\left(Z > -\frac{\mu_{D_t}}{\sigma_{D_t}}\Bigg| \mathcal{H}_{t-1}\right)
    = \mathbb{P}\left(Z < \frac{\mu_{D_t}}{\sigma_{D_t}}\Bigg| \mathcal{H}_{t-1}\right)
    = \Phi \left(\frac{\mu_{D_t}}{\sigma_{D_t}} \Bigg| \mathcal{H}_{t-1}\right),\nonumber\\
    \tilde{\pi}_{t,0}^{\text{TS}} &= \mathbb{P}(\tilde{\mu}_{t,0} > \tilde{\mu}_{t,1} | \mathcal{H}_{t-1}) = \Phi \left(-\frac{\mu_{D_t}}{\sigma_{D_t}} \Bigg| \mathcal{H}_{t-1}\right), \nonumber
\end{align}
where $Z(\cdot)$ denotes the density of a standard normal random variable and $\Phi(\cdot)$ its CDF.

\subsection*{A2. Exact distribution of the TS-based AP-test}

\textbf{General setting.} We first recall that the AP-test for a general $T$-steps experiment is defined as:
\begin{align*}
    AP_{T} = \sum_{t=t_\text{min}}^{T} \mathbb{I}(\pi_{t,1} > 0.5) = 
    \begin{cases}
    0 \quad \text{if}\ \pi_{t,1} \leq 0.5, \quad \forall t\in [1,\dots,T],\\
    1 \quad \text{if}\ \exists_{=1}t \in [1,\dots,T]\!: \pi_{t,1} > 0.5,\\
    2 \quad \text{if}\ \exists_{=2}t \in [1,\dots,T]\!: \pi_{t,1} > 0.5,\\
    \dots\\
    T \quad \pi_{t,1} > 0.5, \quad \forall t\in [1,\dots,T],\\
    \end{cases}
\end{align*}
where $\exists_{=x}$ denotes the existence of exactly $x$ elements that satisfy the condition.

Because of the discrete nature of the AP-test, its exact distribution can be derived by computing the probability that the test takes values $0,\dots, T$ (assuming $t_{\min} = 1$, that will discard $\pi_{0,1}$ which is not informed by the current study's data ).
In practice, \textbf{for doing inference}, it typically suffices to compute $\mathbb{P}(\text{AP}_T > T-1) = \mathbb{P}(\text{AP}_T = T)$ as this is the natural ``extreme'' realisation, generally associated with type-I error and power, as discussed in the main paper. \textbf{In general (i.e., for any randomized adaptive algorithm), and without conditioning on a specific hypothesis (null or alternative)}, such probability can be expressed as:
\begin{align} \label{eq:exact_typeI_general}
    \mathbb{P}(\text{AP}_T = T) &= \mathbb{P}\left(\sum_{t=1}^T\mathbb{I}(\pi_{t,1}>0.5)=T\right) \nonumber\\ 
    &= \mathbb{P}\left(\pi_{1,1}>0.5, \dots,\pi_{T,1}>0.5 \right) \nonumber\\
    &= \mathbb{P}\left(\pi_{1,1}>0.5\right)\prod_{t=2}^T\mathbb{P}\left(\pi_{t,1}>0.5 | \pi_{t-1,1}>0.5, \dots, \pi_{1,1}>0.5\right).
\end{align}

\textbf{Gaussian TS setting.} In particular, under the Gaussian setting as described in Eq.~\eqref{eq:gaussian-setting}, and the TS algorithm, whose exact allocation probabilities are given in Eq.~\eqref{eq: exact_TS}, the closed-form of this probability can be computed by starting decomposing it as:
\begin{align*} 
    \mathbb{P}(\text{AP}^\text{TS}_T
    = T) &= \mathbb{P}\left(\pi^\text{TS}_{1,1}>0.5\right)\prod_{t=2}^T\mathbb{P}\left(\pi^\text{TS}_{t,1}>0.5 | \pi^\text{TS}_{t-1,1}>0.5, \dots, \pi^\text{TS}_{1,1}>0.5\right)\\
    & = \mathbb{P}\left(\Phi \left(\frac{\mu_{D_1}
    }{\sigma_{D_1}
    }\right) > 0.5\right)\prod_{t=2}^T\mathbb{P}\left(\Phi \left(\frac{\mu_{D_t}
    }{\sigma_{D_t}
    }\right) > 0.5 \Bigg| \Phi \left(\frac{\mu_{D_{t-1}}
    }{\sigma_{D_{t-1}}
    }\right) > 0.5,\dots,\Phi \left(\frac{\mu_{D_1}
    }{\sigma_{D_1}
    }\right) > 0.5\right)\\
    & = \mathbb{P}\left(\frac{\mu_{D_1}
    }{\sigma_{D_1}
    } > 0\right)\prod_{t=2}^T\mathbb{P}\left(\frac{\mu_{D_t}
    }{\sigma_{D_t}
    } > 0 \Bigg| \frac{\mu_{D_{t-1}}
    }{\sigma_{D_{t-1}}
    } > 0,\dots,\frac{\mu_{D_{1}}
    }{\sigma_{D_{1}}
    } > 0\right),
\end{align*}
where the last identity comes from the fact that in standard Gaussian distributions $\Phi(x) > 0.5 \iff x>0$.

Now, noticing that, with equal prior distributions (we will be assuming a mean $\mu_\text{prior} = 0$, as considered in general uninformative cases), in the first step $t=0$ we have that $\mathbb{P}(A_{0,i} = 0) = \mathbb{P}(A_{0,i} = 1) = 1/2$ for each $i \in [1,n]$, and taking a fully-sequential setting ($n=1$), by using the law of total probabilities we have that:
\begin{align} \label{eq:prob-T1-1} 
    \mathbb{P}\left(\frac{\mu_{D_1}
    }{\sigma_{D_1}
    } > 0\right) = & \mathbb{P}\left(\frac{\mu_{D_1}
    }{\sigma_{D_1}
    } > 0 \Big| A_{0} = 1\right)\mathbb{P}(A_{0} = 1) + \mathbb{P}\left(\frac{\mu_{D_1}
    }{\sigma_{D_1}
    } > 0 \Big| A_{0} = 0\right)\mathbb{P}(A_{0} = 0) \\
    = & \frac{1}{2}\mathbb{P}\left( \frac{\frac{\sigma^2_\text{prior}Y_{0}(A_0 = 1)}{\sigma^2_y + \sigma^2_\text{prior}}} {\sqrt{\frac{\sigma^2_\text{prior} \sigma^2_y}{\sigma^2_y + \sigma^2_\text{prior}} + \frac{\sigma^2_\text{prior} \sigma^2_y}{\sigma^2_y}}} > 0\right) + \frac{1}{2}\mathbb{P}\left( \frac{- \frac{\sigma^2_\text{prior}Y_{0}(A_0 = 0)}{\sigma^2_y + \sigma^2_\text{prior}}}{\sqrt{\frac{\sigma^2_\text{prior} \sigma^2_y}{\sigma^2_y} + \frac{\sigma^2_\text{prior} \sigma^2_y}{\sigma^2_y + \sigma^2_\text{prior}}}} > 0\right), \nonumber
\end{align}
where we remind that $Y_{t}(A_t = 1) \sim \mathcal{N}(\mu_1, \sigma_y^2)$ and $Y_{t}(A_t = 0) \sim \mathcal{N}(\mu_0, \sigma_y^2)$ for each $t$. Notice that, being a fully-sequential setting, the index $i$ can be omitted, but the same would apply in a batched setting.

Denoting now $c = \frac{\sigma^2_\text{prior}}{\sigma^2_y + \sigma^2_\text{prior}} \Big/ {\sqrt{\frac{\sigma^2_\text{prior} \sigma^2_y}{\sigma^2_y + \sigma^2_\text{prior}} + \frac{\sigma^2_\text{prior} \sigma^2_y}{\sigma^2_y}}}$, with a slight abuse of notation (i.e., denoting with $\mathcal{N}(\cdot, \cdot)$ a random variable with Gaussian distribution), we can see that the above probability can be written as:
\begin{align} \label{eq:prob-T1}
    = & \frac{1}{2}\mathbb{P}\left( \mathcal{N}\left( \mu_1c, \sigma^2_yc^2\right) > 0\right) + \frac{1}{2}\mathbb{P}\left( \mathcal{N}\left( -\mu_0c, \sigma^2_yc^2\right) > 0\right) \nonumber\\
    = & \frac{1}{2}\mathbb{P}\left( \mathcal{N}\left(0, 1\right) > \frac{0 - \mu_1c}{\sigma_yc}\right) + \frac{1}{2}\mathbb{P}\left( \mathcal{N}\left(0, 1\right) > \frac{0 + \mu_0c}{\sigma_yc}\right) \nonumber\\
    = & \frac{1}{2}\left[ \Phi\left( \frac{\mu_1}{\sigma_y}\right) + 1 - \Phi\left( \frac{\mu_0}{\sigma_y}\right) \right].
\end{align}

Alternatively, one could have noticed that $\mathbb{P}(\mathcal{N}\left( \mu_1c, \sigma^2_yc^2\right) > 0) = \mathbb{P}(\mathcal{N}\left( \mu_1, \sigma^2_y\right) > 0)$, with Eq.~\eqref{eq:prob-T1} following directly.

In a two-steps setting (i.e., $t=0,1$), the probability in Eq.~\eqref{eq:prob-T1}, would give us the probability mass of the AP-test statistic distribution correspondent to $T=1$. It is straightforward to notice that under $H_0: \mu_0 = \mu_1$, such probability would be $0.5$, demonstrating the trivial symmetry on the support $\{0,1\}$, i.e., $\mathbb{P}(\text{AP}_1 = 1) = \mathbb{P}(\text{AP}_1 = 0)$ for any $\sigma_y$. On the contrary, under $H_1: \mu_1 > \mu_0$, it is straightforward to verify that $\mathbb{P}(\text{AP}_1 = 1) > \mathbb{P}(\text{AP}_1 = 0)$ regardless the value of $\sigma_y$. Higher the difference between the arms' means ($\mu_1 - \mu_0$), higher the skewness of the AP-test distribution towards the optimal arm. For example, such probability would be $0.6$ in a case with $\mu_0 = 0, \mu_1 = 0.5, \sigma^2_1=1$, corresponding to the power of the test in a $T=1$ setting.

In principle, we have now proved the formula in Eq.~(7) of the main paper. However, one may apply the same logic used in Eq.~\eqref{eq:prob-T1-1}, i.e., the law of total probabilities and properties of the normal distributions, for iteratively expanding and computing the exact probability for a higher number of time steps $T$. 

To illustrate, moving one step further ($T=2$), we start by decomposing the step $t=2$ probability in:
\begin{align} \label{eq:prob-T1-2} 
    \mathbb{P}\left(\frac{\mu_{D_2}
    }{\sigma_{D_2}
    } > 0 \Big | \frac{\mu_{D_1}
    }{\sigma_{D_1}
    } > 0\right) = \frac{\mathbb{P}\left(\frac{\mu_{D_2}
    }{\sigma_{D_2}
    } > 0, \frac{\mu_{D_1}
    }{\sigma_{D_1}
    } > 0\right)}{\mathbb{P}\left(\frac{\mu_{D_1}
    }{\sigma_{D_1}
    } > 0\right)},
\end{align}
where the denominator is known and given in Eq.~\eqref{eq:prob-T1-1}, while the numerator can be decomposed as:
\begin{align*}
    \mathbb{P}\left(\frac{\mu_{D_2}
    }{\sigma_{D_2}
    } > 0, \frac{\mu_{D_1}
    }{\sigma_{D_1}
    } > 0\right) & = \mathbb{P}\left(\frac{\mu_{D_2}
    }{\sigma_{D_2}
    } > 0, \frac{\mu_{D_1}
    }{\sigma_{D_1}
    } > 0 \Big| A_{0}=1, A_{1} = 1\right)\mathbb{P}(A_{0}=1, A_{1} = 1)\\
    & + \mathbb{P}\left(\frac{\mu_{D_2}
    }{\sigma_{D_2}
    } > 0, \frac{\mu_{D_1}
    }{\sigma_{D_1}
    } > 0 \Big| A_{0}=1, A_{1} = 0\right)\mathbb{P}(A_{0}=1, A_{1} = 0)\\
    & + \mathbb{P}\left(\frac{\mu_{D_2}
    }{\sigma_{D_2}
    } > 0, \frac{\mu_{D_1}
    }{\sigma_{D_1}
    } > 0 \Big| A_{0}=0, A_{1} = 1\right)\mathbb{P}(A_{0}=0, A_{1} = 1)\\
    & + \mathbb{P}\left(\frac{\mu_{D_2}
    }{\sigma_{D_2}
    } > 0, \frac{\mu_{D_1}
    }{\sigma_{D_1}
    } > 0 \Big| A_{0}=0, A_{1} = 0\right)\mathbb{P}(A_{0}=0, A_{1} = 0),
\end{align*}
and be solved by computing each separate probability. For example, without loss of generality, focusing on the first addend, we can see that:
\begin{align*}
    \mathbb{P}\left(\frac{\mu_{D_2}
    }{\sigma_{D_2}
    } > 0, \frac{\mu_{D_1}
    }{\sigma_{D_1}
    } > 0 \Big| A_{0}=1, A_{1} = 1\right) & = \mathbb{P} \left( \frac{\frac{\sigma^2_\text{prior}(Y_{0}(A_0 = 1) + Y_{1}(A_1 = 1))}{\sigma^2_y + 2\sigma^2_\text{prior}}} {\sqrt{\frac{\sigma^2_\text{prior} \sigma^2_y}{\sigma^2_y + 2\sigma^2_\text{prior}} + \frac{\sigma^2_\text{prior} \sigma^2_y}{\sigma^2_y}}} > 0, \frac{\frac{\sigma^2_\text{prior}Y_{0}(A_0 = 1)}{\sigma^2_y + \sigma^2_\text{prior}}} {\sqrt{\frac{\sigma^2_\text{prior} \sigma^2_y}{\sigma^2_y + \sigma^2_\text{prior}} + \frac{\sigma^2_\text{prior} \sigma^2_y}{\sigma^2_y}}} > 0 \right)\\
    & = \mathbb{P}\big(Y_{0}(A_0 = 1) + Y_{1}(A_1 = 1) > 0, Y_{0}(A_0 = 1) > 0\big) =\\
    & = \mathbb{P}\big(Y_{0}(A_0 = 1) > 0, Y_{1}(A_1 = 1) > - Y_{0}(A_0 = 1)\big).
\end{align*}

Such probability depends on the underlying parameters of the normal distribution. It can be computed exactly by integration in any case, but in some cases the solution is immediate. For example, in standard Gaussian rewards, since the joint density is radially symmetric, we have that $\mathbb{P}(0<Y_{0}(A_0 = 1)<Y_{1}(A_1 = 1)) = \mathbb{P}(0<Y_{1}(A_1 = 1)<Y_{0}(A_0 = 1)) = \mathbb{P}(0>Y_{0}(A_0 = 1)>Y_{1}(A_1 = 1)) = \mathbb{P}(0>Y_{1}(A_1 = 1)>Y_{0}(A_0 = 1)) = \mathbb{P}(0<-Y_{0}(A_0 = 1)<Y_{1}(A_1 = 1)) = \mathbb{P}(0<Y_{1}(A_1 = 1)<-Y_{0}(A_0 = 1)) = \mathbb{P}(0>-Y_{0}(A_0 = 1)>Y_{1}(A_1 = 1)) = \mathbb{P}(0>Y_{1}(A_1 = 1)>-Y_{0}(A_0 = 1)) = 1/8$. In exactly 3 of these cases, $Y_{0}(A_0 = 1)> 0$ and $Y_{1}(A_1 = 1) > -Y_{0}(A_0 = 1)$, leading to a probability of $3/8$.

\subsection*{A3. Proof of Theorem 3.1 and Lemma 3.2}

\begin{customthm}{Theorem 3.1}\label{eight}
If $\mu_1 > \mu_0$, and arm's allocation probabilities are defined according to standard Thompson Sampling, then the allocation probability $\text{AP}$-test diverges as $T \to \infty$, i.e., $\text{AP}^{\text{TS}}_T \to \infty$.\end{customthm}

\paragraph{\ Proof}
We first recall that the allocation probability AP-test statistic in a $T$-steps experiment is defined as
\begin{align*}
    \text{AP}_T \doteq 
    \sum_{t=1}^T\mathbb{I}\left(\pi_{t,1} > 0.5 \right), 
\end{align*}
with $\pi_{t,1}$ the allocation probability of the experimental arm $1$ at step $t$.

For proving that $\text{AP}_T \to \infty$ as $T \to \infty$ under the alternative ($H_1\!: \mu_1 > \mu_0$), we use the \textit{divergence test}, which provides a sufficient condition for divergence, based on the series's end-behavior, stating that ``if the limit of an $a_t$ series in $\mathbb{R}$ is nonzero, then the sum of the series diverges''. Formally,
\begin{align}\label{eq:div_test}
     \lim_{t \to \infty} a_t \neq 0\quad  \Rightarrow \quad  \sum_{t=n}^\infty a_t= \infty. 
\end{align}

Thus, we only need to show that our sequence $\mathbb{I}(\pi_{t,1} > 0.5)$ has a non-zero limit for $t \to \infty$. As this sequence assumes only values $0$ and $1$, what we need to show is that this limit is $1$. 

By hypothesis, Thompson sampling is used. Using the following  auxiliary result on TS (Lemma 3.2), enables us to demonstrate our thesis. 

\begin{customthm}{Lemma 3.2}
If $\mu_1 > \mu_0$, as the arms are allocated an arbitrarily large number of times, TS converges to allocating the arm $1$ with probability $1$ as $t \to \infty$, i.e.,
\begin{align*}
    \lim_{t \to \infty} \pi_{t,1}^{TS} = 1,\quad \text{when}\  \mu_1 > \mu_0.
\end{align*}
\end{customthm}

\paragraph{\ Proof}
Intuitively, by consistency of posteriors~\citep[see Schwartz theorem, e.g., Theorem 6.16 in][]{ghosal2017fundamentals}, as the number of arm allocations $N_k \to \infty$, $k=0,1$, the expected values of the posterior parameters distributions converge to the true mean rewards $\mu_k$, with $k=0,1$.
Furthermore, the posteriors' variance tends to zero. Thus, if an optimal arm exists, it will gradually and eventually reveal itself, and the probability of allocating that arm by drawing a value $\tilde{\mu}_1 > \tilde{\mu}_0$, according to TS, will tend to unity. As discussed in~\cite{wyatt1998exploration}, if for any finite number of trials, the probability $\pi_{t,k}$ of allocating arm $k$ each trial is greater than $0$ by some finite amount, regardless the history $\mathcal{H}_{t-1}$, the infinite numbers of allocation statement is true. As shown in Eq.~\eqref{eq: exact_TS}, in the case of TS with Gaussian rewards, the allocating probabilities at each finite step $t$ are positive regardless the arm as they are given by the CDF of a normal distribution calculated in a finite value. 

A more rigorous proof of Lemma 3.2, based on the martingale structure inherent in Thompson sampling, can be found in \cite{kalkanli2020asymptotic}. Under the assumption of sub-linear Bayesian regret, authors show that a Thompson sampling policy provides a strongly consistent estimator of the optimal arm. 

Now, according to Lemma 3.2, we have that
\begin{align*}
    \lim_{t \to \infty}\mathbb{I}(\pi_{t,1}^{TS} > 0.5) = \mathbb{I}(\lim_{t \to \infty}\pi_{t,1}^{TS} > 0.5) =  \mathbb{I}(1 > 0.5) = 1.
\end{align*}
By virtue of the divergence test in Eq.~\eqref{eq:div_test}, we have proved our thesis.

\bigskip 

\subsection*{A4. Characteristics and Properties of TS's Allocation Probabilities and TS-based AP-test}
We now give some illustrations of the distribution of the TS allocation probabilities and the related AP-test, providing some insights and properties.

\paragraph{Illustrative examples: relationship between AP-test's distribution and TS's allocation probabilities distribution}

\begin{figure}[h!]
    \centering
    \includegraphics[scale = .62]{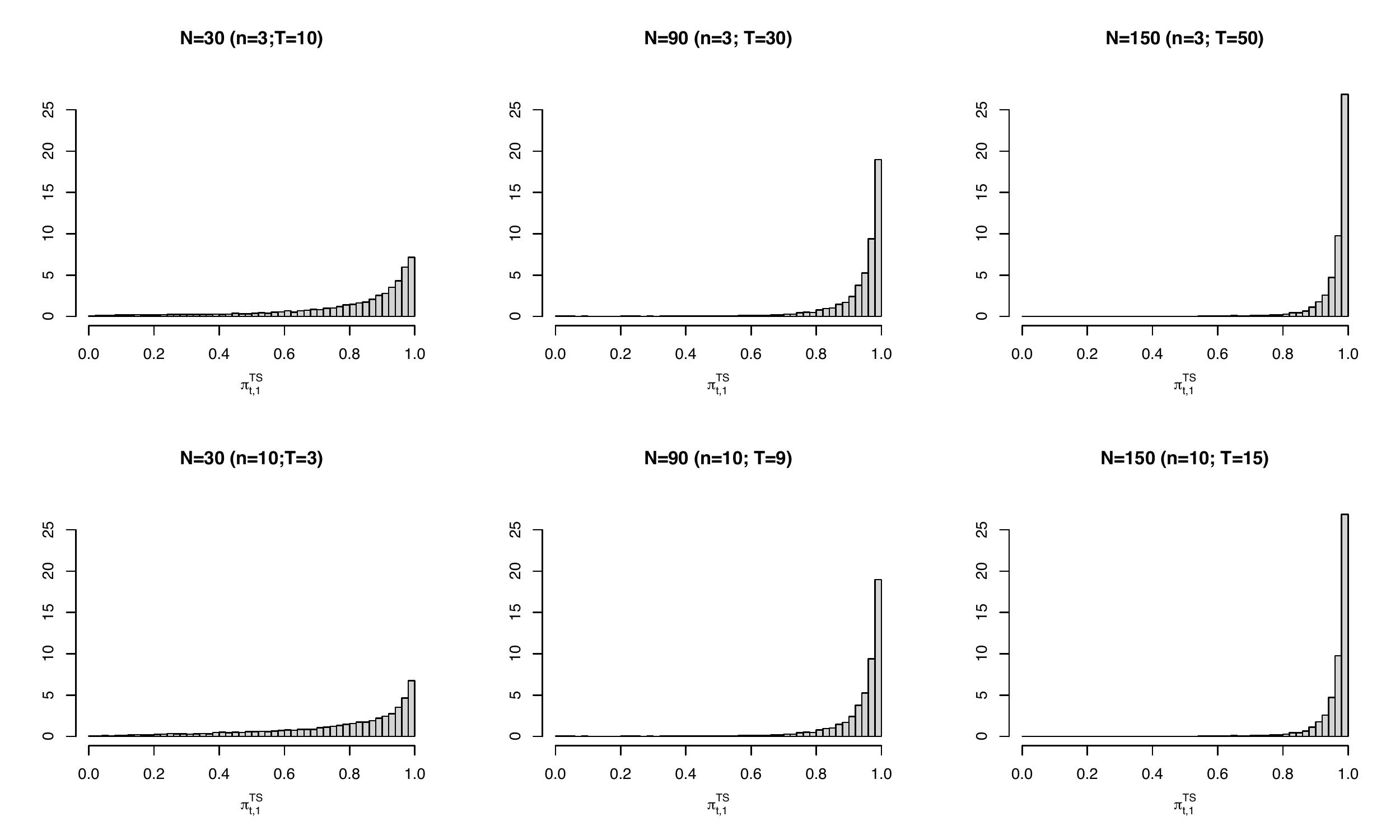}
    \caption{Approximated distribution of TS allocation probability of the experimental arm in the two-arm setting under the alternative hypothesis $H_1: \mu_1>\mu_0$. Reward data generated according to $Y_{t,i} (A_{t,i} = 0) \sim N(0, \sigma^2 = 1)$ and $Y_{t,i} (A_{t,i} = 1) \sim N(0.50, \sigma^2 = 1)$, $t=0,\dots,T$, $i=1,\dots,n$.}
    \label{fig:TS_H1_050}
\end{figure}

\begin{figure}[h!]
    \centering
    \includegraphics[scale = .62]{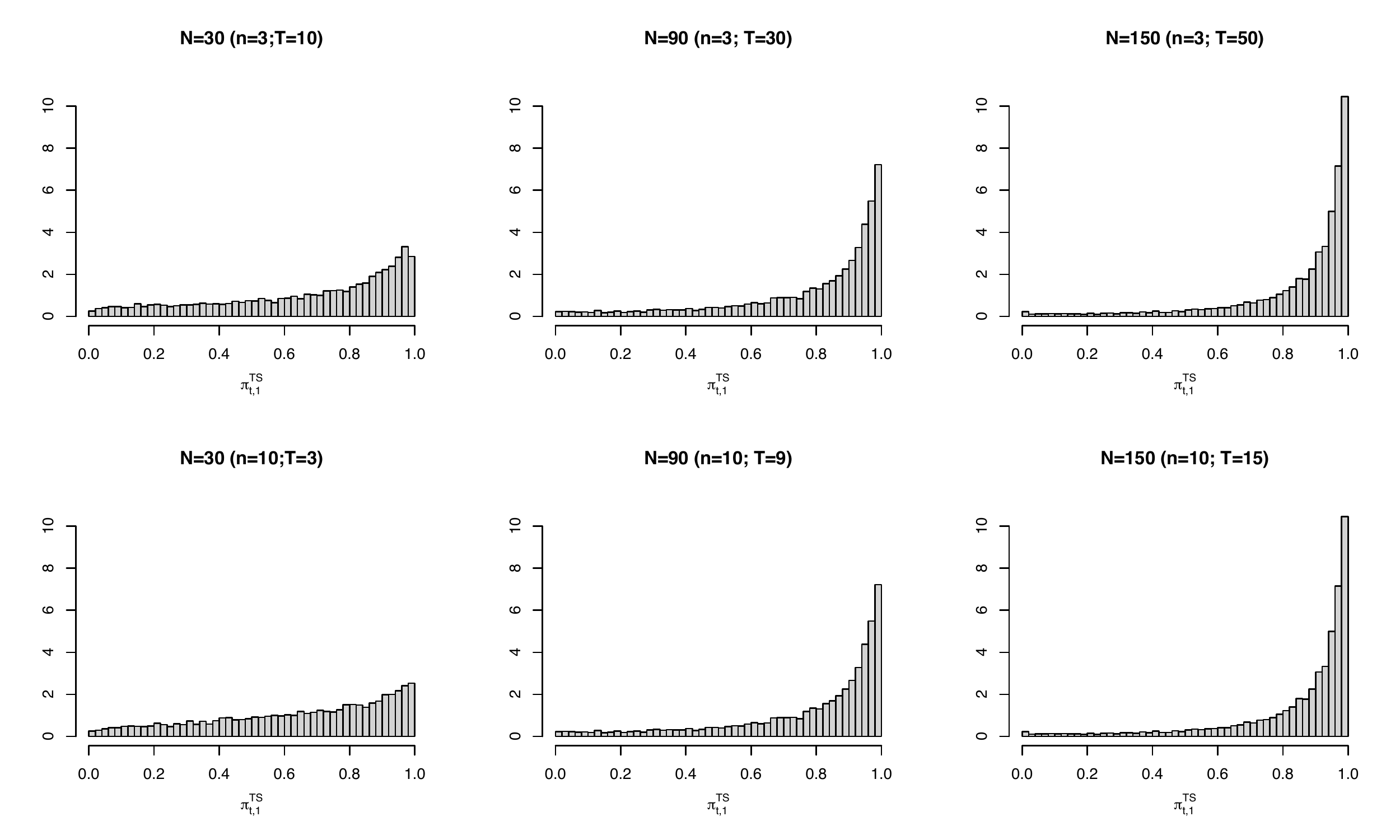}
    \caption{Approximated distribution of TS allocation probability of the experimental arm in the two-arm setting under the alternative hypothesis $H_1: \mu_1>\mu_0$. Reward data generated according to $Y_{t,i} (A_{t,i} = 0) \sim N(0, \sigma^2 = 1)$ and $Y_{t,i} (A_{t,i} = 1) \sim N(0.25, \sigma^2 = 1)$, $t=0,\dots,T$, $i=1,\dots,n$.}
    \label{fig:TS_H1_025}
\end{figure}

As discussed in Section 2.1, \underline{under $H_1$}, the distribution of the TS allocation probabilities is exponentially skewed towards $1$ or $0$, and converges to $1$ for the optimal arm and to $0$ for the sub-optimal arm(s) as $t \to \infty$. The convergence rate to the extremes does not, or only negligibly (at the beginning) depend on the update frequency, i.e., batch size, but it substantially depends on the underlying true differences between the arm means (see the comparison between Figures~\ref{fig:TS_H1_050} and \ref{fig:TS_H1_025}).  
This justifies the choice of the alternative AP-test statistic we proposed here, which directly depends on the allocation probabilities of the design, as these may be able to detect cases in which there's no underlying difference between the arm means ($H_0$), and cases in which there exists a difference ($H_1$). In addition, higher this difference, greater the ability of the test statistic to rapidly (i.e., with a lower sample size $N$) detect it. 

\begin{figure}[h!]
    \centering
    \includegraphics[scale = .65]{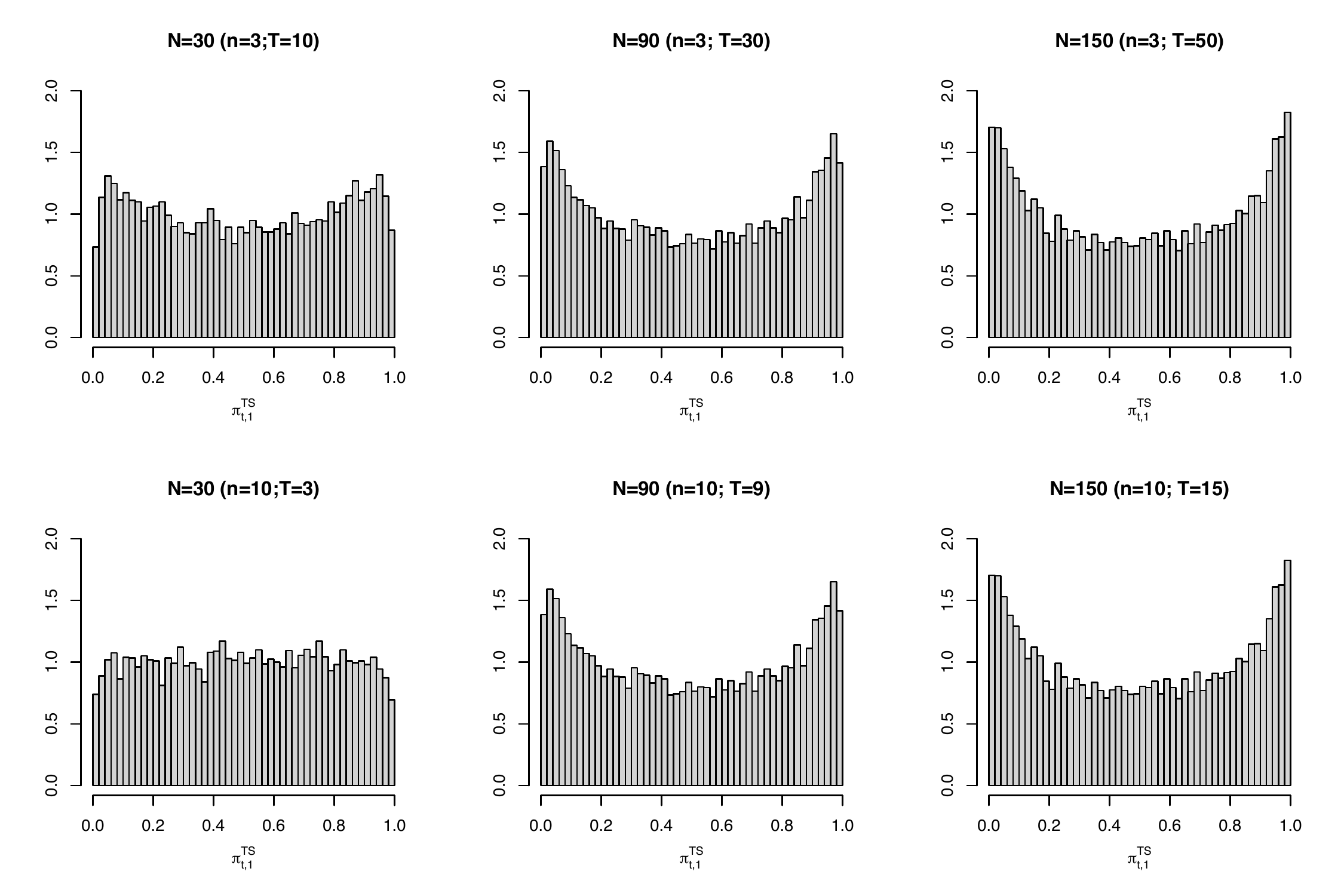}
    \caption{Approximated distribution of TS allocation probability of the experimental arm in the two-arm setting under the null hypothesis $H_0: \mu_0=\mu_1$. Reward data generated according to $Y_{t,i} (A_{t,i}=k) \sim N(0, \sigma^2 = 1)$, $k=0,1$, $t=0,\dots,T$, $i=1,\dots,n$.}
    \label{fig:TS_H0}
\end{figure}

\underline{Under $H_0$}, the distribution of interest and its asymptotic behavior has a very different pattern. It has again a support in $[0,1]$ a symmetry around $0.5$, but it seems to converge to a U-shape distribution with peaks closer to the extremes as $t \to \infty$ (see Figure~\ref{fig:TS_H0}). Asymptotically, the TS allocation probabilities do not concentrate, and in the specific case of step $t=1$, it has a uniform distribution in $[0,1]$~\citep{Zhang_NEURIPS2020}.
\begin{figure}[h!]
    \centering
    \includegraphics[scale = .57]{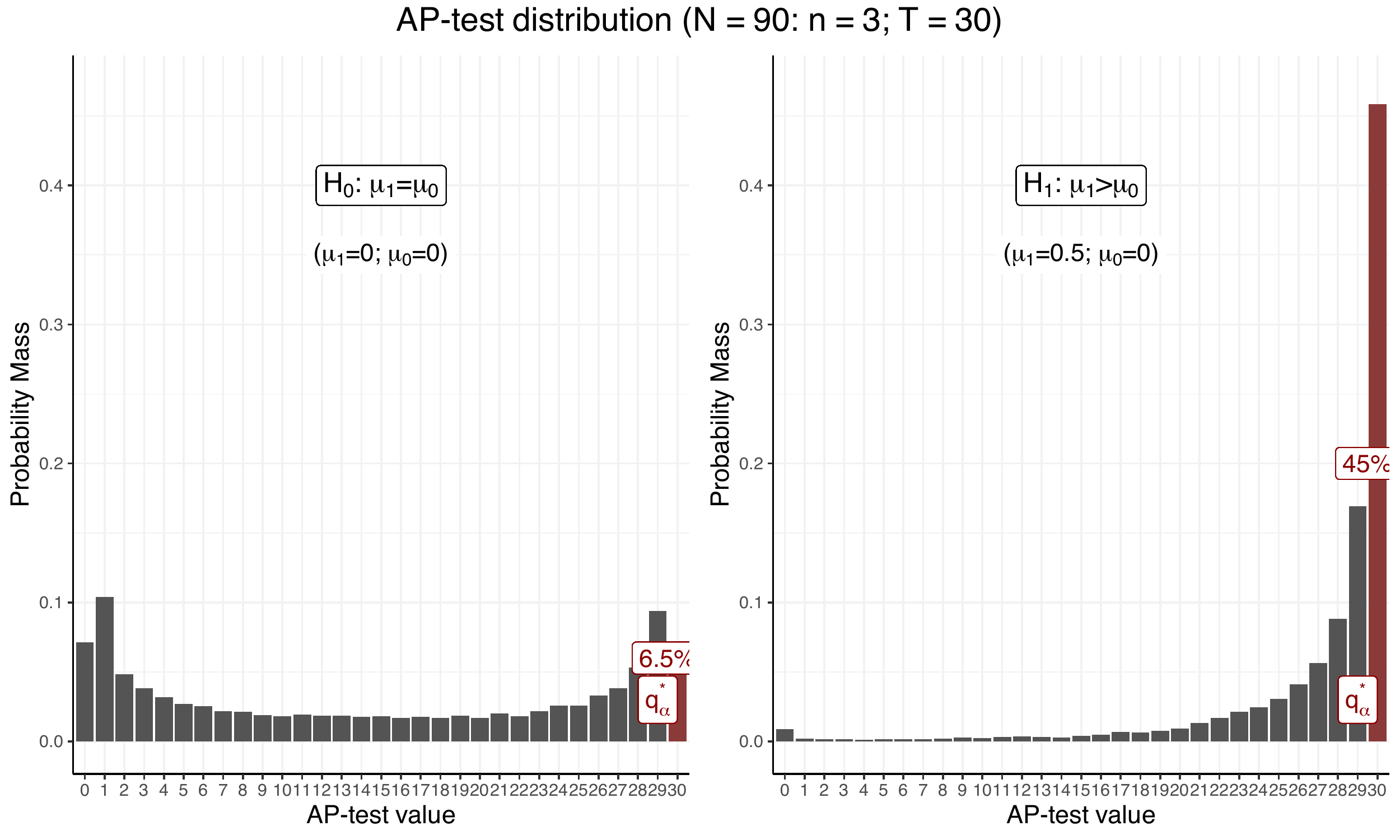}
    \caption{Distribution of the AP-test in a two-arm setting under $H_0\!: \mu_0 =  \mu_1$  and $H_1\!: \mu_1 > \mu_0$  Reward data generated according to $Y_{t,i} (A_{t,i} = 0) \sim N(0, \sigma^2 = 1)$ and $Y_{t,i} (A_{t,i} = 1) \sim N(0.5, \sigma^2 = 1)$, $t=0,\dots,T$, $i=1,\dots,n$.
    }
    \label{fig: Pi_H0_H1}
\end{figure}

The pattern of the TS allocation probabilities distribution is very well reflected in the distribution of the AP-test. As illustrated in Figure~\ref{fig: Pi_H0_H1}, it the same symmetric distribution (now around the midpoint of $T/2$) with peaks around the extremes under $H_0$ and with a very similar shape: more specifically, compare top-middle plot of Figure~\ref{fig:TS_H0} and the left plot of Figure~\ref{fig: Pi_H0_H1}. Under the alternative, it is a skewed distribution favouring the optimal arm: more specifically, compare top-middle plot of Figure~\ref{fig:TS_H1_050} and the right plot of Figure~\ref{fig: Pi_H0_H1}. 

However, by definition, the distribution of the AP-test is a discrete distribution with support $\{0,1,\dots, T\}$, and thus changes with $T$. Notably, the asymptotic convergence properties of TS, and its different convergence rate to the optimal arm under $H_0$ and $H_1$, ensures, for a given $T$, systematically lower mass in correspondence to the critical value $q_\alpha^*$ under the null (i.e., type-I error), compared to the analogous mass under the alternative (i.e., statistical power). 

\paragraph{Properties of the AP-test (1): Type-I error is a decreasing function of step $\boldsymbol{t}$}

Here, we show that type-I error is a decreasing function of $t$, allowing us to continuously improve upon this error while collecting more data. 

We start by recalling that, according to Eq. (6) of the main paper that defines the critical value $q^*_\alpha$, the AP-test's type-I error is:
\begin{align*}
    \mathbb{P}(\text{AP}_T > q^*_\alpha | H_0) = \mathbb{P}(\sum_{t=1}^T\mathbb{I}\left(\pi_{t,1} > 0.5 \right) > q^*_\alpha | H_0).
\end{align*}
We remind that, by definition, the distribution of $\text{AP}_T$ is a discrete distribution with support $\{0,\dots,T \}$. It thus depends on the experimental size or horizon $T$. This also applies to the critical value $q^*_\alpha$, which in general (for small or moderate $T$'s) is equal to $T-1$ as illustrated in Figure~\ref{fig: Pi_H0_H1}. Assuming then $q^*_\alpha = T-1$, and taking into account that the maximum value of $\text{AP}_T$ is $T$, we have that
\begin{align}\label{eq:typeI_decre}
     \mathbb{P}(\text{AP}_T \geq q^*_\alpha | H_0) =&\mathbb{P}\left(\sum_{t=1}^T\mathbb{I}\left(\pi_{t,1} > 0.5 \right) \geq T | H_0\right) \nonumber\\
     =& \mathbb{P}\left(\sum_{t=1}^T\mathbb{I}\left(\pi_{t,1} > 0.5 \right) = T | H_0\right) \nonumber\\
     =& \mathbb{P}\left(\mathbb{I}\left(\pi_{1,1} > 0.5 \right) = 1, \dots, \mathbb{I}\left(\pi_{T,1} > 0.5 \right) = 1 | H_0\right) \nonumber\\
     =& \mathbb{P}\left(\pi_{1,1} > 0.5, \dots, \pi_{T,1} > 0.5 | H_0\right) \nonumber\\
     =& \mathbb{P}(\pi_{1,1} > 0.5 | H_0) \times \mathbb{P}(\pi_{2,1} > 0.5| H_0, \pi_{1,1} > 0.5)\times \cdots \nonumber\\ &\cdots \times \mathbb{P}(\pi_{T,1} > 0.5| H_0, \pi_{1,1} > 0.5, \dots, \pi_{(T-1),1} > 0.5).
\end{align}

Note that each of the factors in Eq.~\eqref{eq:typeI_decre} is a positive value in $(0,1)$; in the specific case of Gaussian rewards they are given in Eq.~\eqref{eq: exact_TS}. This means that, as $t$ increases, the number of factors increases, decreasing thus total probability, i.e., the overall the type-I error. The rate at which this probability decreases depends on the conditioning event or hypothesis. We expect it to converge to $0$ under $H_0$, and to some fixed value under $H_1$, with a convergence rate depending on the exact specifications of $H_0$ and $H_1$. 

\paragraph{Properties of the AP-test (2): Power is not affected by constraints on algorithm's allocation probabilities, or probability clipping.}

\begin{figure}[h]
    \centering
    \includegraphics[scale = .6]{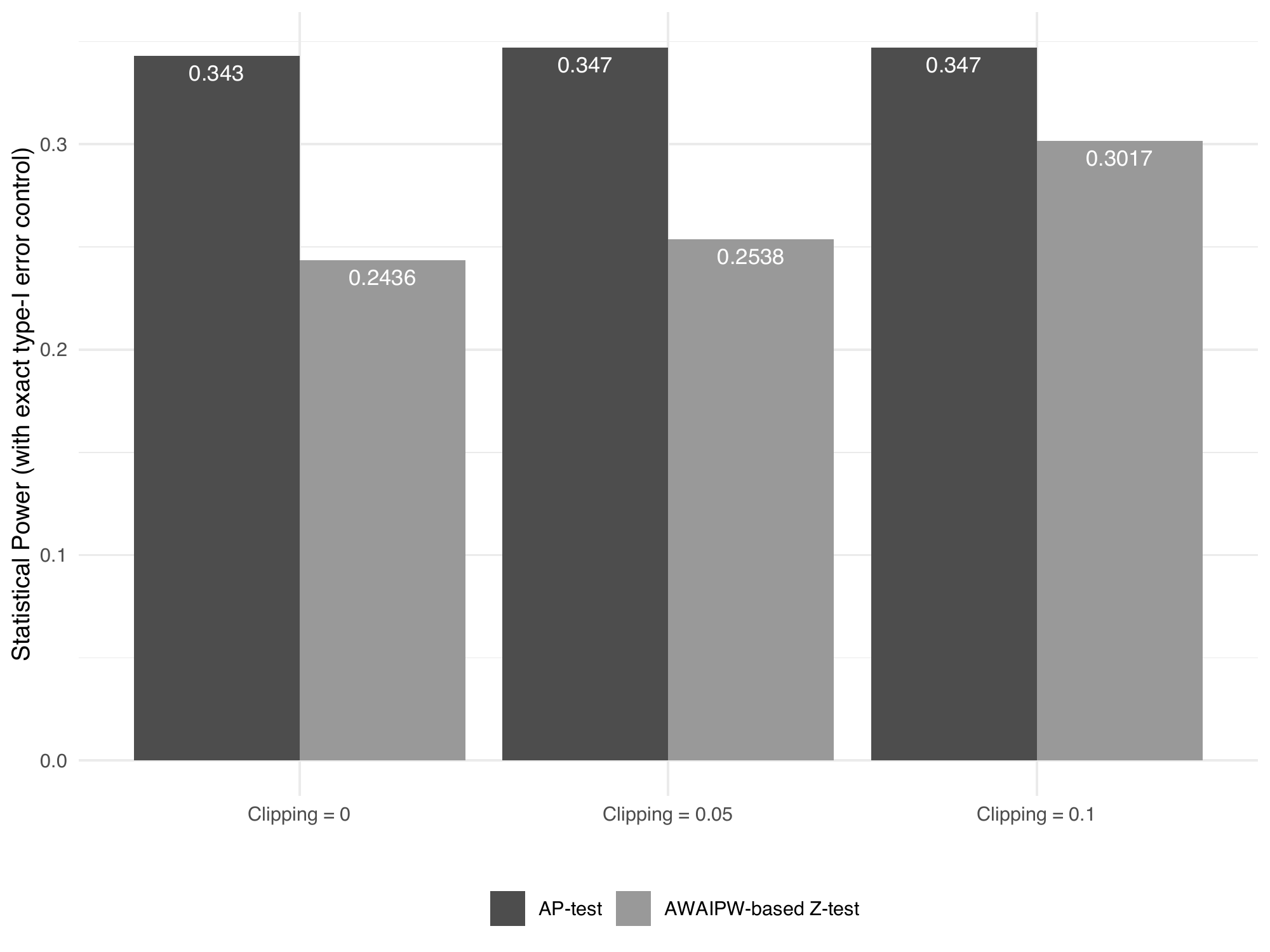}
    \caption{Power (with an $\alpha = 0.05$ exact type-I error control) of the AP-test, compared to the AW-AIPW approach, for different probability clipping values (i.e., $\{0, 0.05, 0.1\}$). Reward data simulated according to TS in a setting analogous to the real-world deployment in Section 5: $N=150$, $T=50$, $n=3$. A total of $10000$ MC trajectories were considered, with standard errors always $<0.002$.
    }
    \label{fig:power_robust_clip}
\end{figure}
Literature suggests that in order to perform valid inference on bandit data using traditional inferential methods it is necessary to guarantee that the bandit algorithm explores sufficiently. For example, existing estimators proposed in the adaptive setting, such as the BOLS~\citep{Zhang_NEURIPS2020} and the AW-AIPW~\citep{hadad2021confidence}, are based on CLTs with conditions that implicitly impose constraints on the actions' sampling probabilities of the bandit algorithm (see Definition 1 in~\cite{Zhang_NEURIPS2020} and Assumptions 1-3 in~\cite{hadad2021confidence}). These include avoiding action selection probabilities below certain thresholds (i.e, probability clipping) and/or consideration on the rate of convergence of these probabilities (see also Supplementary material C3). 
In addition, other works~\citep[see e.g.,][]{yao_power} also discuss how greater exploration with probability clipping may solve the problem of reduced power of a statistical tests in adaptively-collected data. Such property is supported by Figure~\ref{fig:power_robust_clip}, where the statistical power of the AW-AIPW-based Z-test is shown to improve by increasing the clipping threshold. 

However, limiting the exploitative behaviour of a bandit algorithm may have substantial regret trade-offs (as illustrated in Figure 1 in the main paper), especially when the optimal arm exists and has a remarkable benefit compared to sub-optimal ones. 

\textbf{Differently form existing work, our AP-test: 1) does not require constraining the algorithm with probability clipping for ensuring valid inference, and 2) this procedure is not associated with an improved power. Alternatively said: a non-restricted algorithm will not translate in a reduced power}. As illustrated in Figure~\ref{fig:power_robust_clip}, with a simulation setting analogous to our real-world experiment in Section 5 (main paper), power remains quite constant and thus robust with respect to the probability clipping rate of the bandit procedure, while this is not verified for the AW-AIPW procedure.

\newpage
\section*{B. Randomization Procedure for Exact Type-I Error Control}

As extensively discussed in the main paper, when the distribution of test statistic is discrete under $H_0$, it might not be possible to choose the critical value $q_\alpha^*$ so that the significance level is exactly $\alpha$. While this is a characteristic of discrete tests, reason why certain tricks have been introduced for achieving exact $\alpha$ control (see e.g., \url{https://web.stanford.edu/class/archive/stats/stats200/stats200.1172/Lecture06.pdf}), note that continuous tests may also suffer from type-I error inflation, especially in small samples, and this issue is typically under-reported in literature. Thus, we extend the discrete tests procedure to a continuous setting as well. 

We illustrate here an approach that can be used for achieving exact type-I error control with any given level $\alpha$, in both discrete and continuous tests. In addition to allowing researchers to perform inference in a more reliable (i.e., robust) way, such procedure will guarantee a high comparability between methods, when e.g., power is of interest.

In what follows, we assume the same superiority hypothesis testing problem considered in the main paper. 

\subsection*{B1. Discrete Tests}

\textbf{With reference to the example in Figure 2 (main paper).} Let's consider the example reported in Figure 2 (left plot–under $H_0$) of the main paper. Suppose that, rather than considering an $\alpha = 0.1$, we wish to achieve significance level $\alpha = 0.05$.  
In our example, based on a setting with sample size $N=51$, a number of $T=17$ steps, and a batch size of $n=3$, we have that the type-I error of the $\text{AP}_T$ test, according to the originally specified critical value $q_{\alpha = 0.1}^* = 16$, is $7.3\%$. The only other value that would allow for a higher control (i.e., $<7.3\%$), is the most extreme value that can have the test statistic i.e., $17$. However this will lead to a very conservative control of $0\%$, that will translate in a $0\%$ power. Here we illustrate a strategy to avoid such issue and achieve an exact $\alpha = 0.05$, control.

More formally, we have that:
\begin{align*}
    \mathbb{P}(\text{AP}_T > 16 | H_0) = \mathbb{P}(\text{AP}_T = 17 | H_0) = 0.073\quad &\Rightarrow \quad \text{we do not achieve a significance level} \leq \alpha = 0.05\\
    \mathbb{P}(\text{AP}_T > 17 | H_0) = \mathbb{P}(\text{AP}_T = 18 | H_0) = 0\quad &\Rightarrow \quad \text{we are too conservative compared to}\ \alpha = 0.05.
\end{align*}
The  theoretically  correct  solution for achieving an intermediate error probability of $0.05$ is  to  perform  a  randomized  test, that proceeds as follows.
\begin{align*}
    \text{If}\ \text{AP}^\text{observed}_T \leq 16\quad &\Rightarrow \quad \text{Do not reject}\ H_0 \\
    \text{If}\ \text{AP}^\text{observed}_T = 17\quad &\Rightarrow \quad \text{Reject}\ H_0\ \text{with probability}\ \gamma, 
\end{align*}
where $\gamma$ is chosen to make the significance level exactly $\alpha$. In our case this is given by $0.68 = 0.05/0.073$, meaning that, for achieving exact $0.05$-level control, we will randomly reject $H_0$ in $68\%$ of the case where we have an observed test $\text{AP}^\text{observed}_T = 17$.

\textbf{General case.} Let's now extend the procedure to a general case where we may also have values that lead to a conservative ( $< 0.05$), but non-null type-I error, i.e., 
\begin{align*}
    \mathbb{P}(\text{Test} > q_t | H_0) &= \overline{\alpha}_t > \alpha = 0.05,\\
    \mathbb{P}(\text{Test} > q_{t+1} | H_0) &= \overline{\alpha}_{t+1} \in [0, \alpha = 0.05],
\end{align*}
where $q_t < q_{t+1}$, and $\overline{\alpha}_t > 0.05 \geq \overline{\alpha}_{t+1} \geq 0$. 

According to the previous logic, one would apply the randomized test as follows:
\begin{align*}
    \text{If}\ \text{Test}^\text{observed} \leq q_t\quad &\Rightarrow \quad \text{Do not reject}\ H_0 \\
    \text{If}\ \text{Test}^\text{observed} > q_{t+1}\quad &\Rightarrow \quad \text{Reject}\ H_0 \\
    \text{If}\ \text{Test}^\text{observed} = q_{t+1}\quad &\Rightarrow \quad \text{Reject}\ H_0\ \text{with probability}\ \gamma \doteq \frac{\alpha - \overline{\alpha}_{t+1}}{\overline{\alpha}_{t} - \overline{\alpha}_{t+1}}, 
\end{align*}
where one may notice that $\overline{\alpha}_{t} - \overline{\alpha}_{t+1} = \mathbb{P}(\text{Test} = q_{t+1} | H_0)$.

It is straightforward to verify that this procedure leads to an exact type-I error control as:
\begin{align*}
    \text{Type-I Error} &\doteq \mathbb{P}(\text{Reject}\ H_0 | H_0)\\ &= \mathbb{P}(\text{Test} > q_{t+1} | H_0) + \gamma \mathbb{P}(\text{Test} = q_{t+1} | H_0) = \overline{\alpha}_{t+1} + \frac{\alpha - \overline{\alpha}_{t+1}}{\overline{\alpha}_{t} - \overline{\alpha}_{t+1}}(\overline{\alpha}_{t} - \overline{\alpha}_{t+1}) = \alpha.
\end{align*}

\subsection*{B2. Continuous Tests}

For achieving an exact type-I error control in continuous tests, we propose an extension of the same randomized strategy used in discrete tests, with the difference that in the former we are not able to compute the probability of a test being equal to a certain threshold value $q_{t+1}$, e.g., $\mathbb{P}(\text{Test} = q_{t+1} | H_0)$, given that $q_{t+1} \in \mathbb{R}$ or a continuous subset of it. To the best of our knowledge, the continuous variant is a new extension of the method for discrete tests.

\textbf{General case.} We define the problem as follows. Let's consider a continuous Test, e.g., a Z-test with standard Gaussian distribution under the null, or any another test defined on a subset of $\mathbb{R}$ with a known theoretical distribution. Despite the knowledge of the exact critical value, say $t^*$, that should ensure type-I error control, due to the asymptotic nature of these tests, such control may not be verified in finite (especially small) samples, i.e.,  
\begin{align*}
    \mathbb{P}(\text{Test} > t_\alpha^* | H_0) &= \overline{\alpha} > \alpha = 0.05.
\end{align*}

Using the same logic as in Section B1, we proceed as follows:
\begin{align*}
    \text{If}\ \text{Test}^\text{observed} < t_\alpha^* \quad &\Rightarrow \quad \text{Do not reject}\ H_0 \\
    \text{If}\ \text{Test}^\text{observed} > t_\alpha^*\quad &\Rightarrow \quad \text{Reject}\ H_0\ \text{with probability}\ \gamma \doteq \frac{\alpha}{\overline{\alpha}}.
\end{align*}

\textbf{Remark}. Clearly, any type-I error adjustment, made in both discrete and continuous tests, needs to be included in the power calculations, where the rejection rate will be properly scaled according to probability $\gamma$. Thus, when we adjust for a type-I error inflation we will see an (adjusted) power reduction, while this will increase in case of conservative tests and significance levels below the pre-specified threshold $\alpha$.

\newpage 

\section*{C. Comparative Testing Approaches}
Here, we report the testing approaches we used in our two-arm setting simulated experiments for hypothesis testing performance assessment. 

\subsection*{C1. Batched Ordinary Least Square (BOLS)-based Z-test}
The idea of the BOLS estimator is to perform standard Ordinary Least Squared (OLS) estimation in each step separately, and then combine the individual information with weights defined by the individual steps' variance.

We remind that the general OLS estimator is given by: 
\begin{align}
    \hat{\boldsymbol{\mu}}^{\text{OLS}} = (\mathbf{X}^T\mathbf{X})^{-1}\mathbf{X}^T\mathbf{Y},
\end{align}
where $\mathbf{X} \doteq [\mathbf{X}_{0,1},\dots,\mathbf{X}_{0,n},\dots,\mathbf{X}_{T,1},\dots,\mathbf{X}_{T,n}]^T \in \mathbb{R}^{nT \times 2}$ denotes the design matrix, and $\mathbf{Y} \doteq [\mathbf{Y}_{0,1},\dots,\mathbf{Y}_{0,n},\dots,\mathbf{Y}_{T,1},\dots,\mathbf{Y}_{T,n}]^T \in \mathbb{R}^{nT}$ the response variables. 

In a two-arm non-contextual setting, in each step $t$ and for each participant $i$, the design matrix reduces to a vector of dummies with length equal to the number of arms, where value $1$ indicates that that arm was selected. More formally, we have that, in each step $t$ and for each participant $i$, $X_{t,i} = [1-A_{t,i}, A_{t,i}]$. Note that $\mathbf{X}^T\mathbf{X} \doteq \sum_{t=0}^T\sum_{i=1}^nX_{t,i}X_{t,i}^T$.

Assuming Gaussian rewards, the OLS of each of the unknown parameters is equivalent to their Maximum Likelihood Estimator (MLE), i.e.,
\begin{align}
    \hat{\mu}_0^{\text{OLS}} = \frac{\sum_{t=0}^T\sum_{i=1}^n (1-A_{t,i})Y_{t,i}}{N_0},\quad \hat{\mu}_1^{\text{OLS}} = \frac{\sum_{t=0}^T\sum_{i=1}^n A_{t,i}Y_{t,i}}{N_1},
\end{align}
where $N_0 \doteq \sum_{t=0}^T\sum_{i=1}^n (1-A_{t,i})$ and $N_1 \doteq \sum_{t=0}^T\sum_{i=1}^n A_{t,i}$ represent the number of times arm $0$ and arm $1$ were allocated during the experiment, respectively. 

Denoted now with $\Delta \doteq \mu_1 - \mu_0$ our quantity of interest, its standard OLS estimator is given by $\hat{\Delta}^{\text{OLS}} \doteq \hat{\mu}_1^{\text{OLS}} - \hat{\mu}_0^{\text{OLS}}$. Notably, such estimator leads to invalid inference in the adaptive setting as demonstrated in~\cite{Zhang_NEURIPS2020}. A valid alternative, with asymptotic theoretical guarantees is then proposed as follows. 

Focusing on individual step $t$ differences, one could first estimate each $\Delta_t \doteq \mu_{t,1} - \mu_{t,0}$ separately, for all $t$'s, by using individual OLS estimators (named batched OLS or BOLS), given by $\hat{\Delta}_t^{\text{BOLS}} \doteq \hat{\mu}_{t,1}^{\text{BOLS}} - \hat{\mu}_{t,0}^{\text{BOLS}}$, with 
\begin{align}
    \hat{\mu}_{t,0}^{\text{BOLS}} = \frac{\sum_{i=1}^n (1-A_{t,i})Y_{t,i}}{N_{t,0}},\quad \hat{\mu}_{t,1}^{\text{BOLS}} = \frac{\sum_{i=1}^n A_{t,i}Y_{t,i}}{N_{t,1}},
\end{align}

and then combine them to obtain the BOLS-based Z-test statistic,  defined as:
\begin{align*}
    Z_{\text{BOLS}} = \frac{1}{T}\sum_{t=0}^T\frac{\sqrt{N_{t,0}N_{t,1}}\left(\hat{\Delta}_t^{\text{OLS}} - \Delta_t\right)}{\sqrt{(N_{t,0}+N_{t,1})\hat{\sigma}_t^2}},
\end{align*}
where $N_{t,0} \doteq \sum_{i=1}^n (1-A_{t,i})$ and $N_{t,1} \doteq \sum_{i=1}^n A_{t,i}$ represent the number of times arm $0$ and arm $1$ were allocated up to step $t$, respectively.

In such a case the noise variance is estimated for each step
$t$ as follows:
\begin{align*}
    \hat{\sigma}_t^2 = \frac{1}{n-2}\sum_{i=1}^n\left(Y_{t,i} - (1-A_{t,i})\hat{\mu}_{t,0}^{\text{BOLS}} - A_{t,i}\hat{\mu}_{t,1}^{\text{BOLS}} \right)^2,
\end{align*}
with $n-2$ being the overall degrees of freedom related to one single batch. Consistency of $\hat{\sigma}_t^2$ is provided in~\cite{Zhang_NEURIPS2020}.

Since we are in an unknown variance setting, for computing the critical values $t_\alpha^*$ required for performing inference, we use the Student-$t$ distribution (accounting for the batched combination), rather the the standard normal one, typically used in known variance settings. 

More specifically, given $\alpha$, we defined $t_\alpha^*$ as
\begin{align*}
    \mathbb{P}\left( Z_{\text{BOLS}} \geq t_\alpha^*\right) = \mathbb{P}\left( \sum_{i=1}^T Y_{t} \geq t_\alpha^*\right) = \alpha,
\end{align*}
with $Y_{t} \sim t_{n-2}$ iid random variables, and reject the null hypothesis if $Z_{\text{BOLS}} \geq t_\alpha^*$.

We note that such sum of iid Student-$t$ distribution is unknown. Thus, the critical value is derived according to the estimated Monte Carlo distribution for each step or batch $t$ as in~\cite{Zhang_NEURIPS2020}.

\subsection*{C2. Adaptively Weighted Augmented Inverse Probability Weighting (AW-AIPW)-based Z-test}

Since multiple versions of the AW-AIPW estimator for treatment differences are given in different works, we report here the original version available in the Supplementary material of~\cite{hadad2021confidence}.

Denote again with $\Delta \doteq \mu_{1} - \mu_{0}$ the quantity of interest and $\hat{\Delta}^{\text{AW-AIPW}} \doteq \hat{\mu}_{1}^{\text{AW-AIPW}} - \hat{\mu}_{0}^{\text{AW-AIPW}}$ its AW-AIPW estimator. The AW-AIPW idea is to compute the Augmented Inverse Probability Weighting (AIPW) estimator at each step $t$ and to weight it according to certain \textit{adaptive weights} $h_t$. 

More specifically, defining the AIPW estimator of arm $0$ and arm $1$ at each step $t$ and after each new observation (or participant) $i$ as:
\begin{align*}
    \hat{\Gamma}_{t,i,0} \doteq \frac{(1-A_{t,i})Y_{t,i}}{\pi_{t,0}} + \left( 1- \frac{1-A_{t,i}}{\pi_{t,0}}\right)\hat{m}_{t,0};\quad \quad 
    \hat{\Gamma}_{t,i,1} &\doteq \frac{A_{t,i}Y_{t,i}}{\pi_{t,1}} + \left( 1- \frac{A_{t,i}}{\pi_{t,1}}\right)\hat{m}_{t,1},
\end{align*}
with $\hat{m}_{t,0}$ and $\hat{m}_{t,1}$ a step $t$ estimate of the conditional mean reward for arm $0$ and $1$, respectively, and $\pi_{t,\cdot}$ the allocation probabilities at each step $t$, the $\hat{\mu}_{1}^{\text{AW-AIPW}}$ and $\hat{\mu}_{0}^{\text{AW-AIPW}}$ estimators are given by:
\begin{align*}
    \hat{\mu}_{0}^{\text{AW-AIPW}} = \sum_{t=0}^T\sum_{i=1}^n\left(\frac{h_{t,0}}{\sum_{t=0}^T\sum_{i=1}^nh_{t,0}}\right)\hat{\Gamma}_{t,i,0};\quad \quad 
    \hat{\mu}_{1}^{\text{AW-AIPW}} = \sum_{t=0}^T\sum_{i=1}^n\left(\frac{h_{t,1}}{\sum_{t=0}^T\sum_{i=1}^nh_{t,1}}\right)\hat{\Gamma}_{t,i,1}.
\end{align*}
We take $\hat{m}_{t,k}$ to be the sample mean of each arm $k$ up to step $t-1$ (as in the original paper), i.e., 
\begin{align*}
    \hat{m}_{t,k} = \frac{\sum_{j = 0}^{t-1}A_{t,i}Y_{t,i}}{\sum_{j = 0}^{t-1}A_{t,i}}, \quad t=0,\dots, T, k=0,1,
\end{align*}
and we use the variance stabilizing weights, equal to the square root of the allocation probabilities, i.e.,
\begin{align*}
    h_{t,k} = \sqrt{\pi_{t,k}}, \quad t=0,\dots, T, k=0,1.
\end{align*}

Now, for hypothesis testing, the following studentized statistic, proved to be asymptotically standard normal, is proposed:
\begin{align*}
    Z_{\text{AW-AIPW}} =  \frac{\hat{\Delta}^{\text{AW-AIPW}} - \Delta}{\sqrt{\hat{V}_1 + \hat{V}_2}},
\end{align*}
where each $\hat{V}_k$, for $k=0,1$, is defined as:
\begin{align*}
    \hat{V}_k \doteq \frac{\sum_{t=0}^T\sum_{i=1}^nh_{t,k}^2\left(\hat{\Gamma}_{t,i,k} - \hat{\mu}_{k}^{\text{AW-AIPW}} \right)^2}{\left(\sum_{t=0}^T\sum_{i=1}^nh_{t,k}\right)^2}.
\end{align*}

Note that, differently from the BOLS procedure of~\cite{Zhang_NEURIPS2020}, which is computed in batches, the AW-AIPW approach is computed recursively after each new observation $i$ in each step $t$. In a general batched setting this means that we perform an estimation at each point $t' = 0,\dots, nT$. \textbf{Such requirement implies a computational cost much higher compared to both BOLS-based Z-test and the AP-test}. Note that this is not only because of the need of computing the AW-AIPW for each $t' \in [0, nT]$ (compared to the $t \in [0, T]$ of the other approaches), but also because the computation of the AW-AIPW estimator involves the estimation of the augmented reward model part (i.e., $\hat{m}_{t,\cdot}$), and the adaptive weights ($h_{t,\cdot}$), in addition to the IPW estimator.

For ensuring comparability between methods, since our proposed strategy discards the first batch (i.e., $t=0$) as this is associated with an equal allocation probability (when identical priors are considered for both arms), inference will be performed starting from $t=1$ for each strategy. 

\subsection*{C3. Restricted (BOLS and AW-AIPW) Thompson Sampling}

As noted in the main paper, the comparative approaches based on the BOLS and the AW-AIPW estimators assume certain restrictions on the adaptive algorithm. In general, first tentative assignment
probabilities are computed via Thompson sampling, and then they are adjusted to impose lower and upper bounds as follows:
\begin{description}
 \item[Restricted TS (BOLS)] Allocation probabilities of the TS are restricted to belong in the range $[\pi_{\text{min}}, \pi_{\text{max}}]$, meaning that no allocation probabilities lower than $\pi_{\text{min}}$ or higher than $\pi_{\text{max}}$ are allowed. Formally, using the thresholds proposed in~\cite{Zhang_NEURIPS2020}, TS's allocation probabilities $\tilde{\pi}^{\text{TS}}_{t,k}$ are redefined (clipped) as:
 \begin{align*}
    \pi_{\text{min}} &=  0.1; \quad \pi_{\text{max}} = 0.9\\
     \pi^{\text{TS-BOLS}}_{t,k} &= \min \left(\pi_{\text{max}},  \max \left(\tilde{\pi}^{\text{TS}}_{t,k}, \pi_{\text{min}}\right)\right), \quad \forall t \in [0,T], k=0,1.
 \end{align*}
 In addition, in order to apply this batched estimation approach, it should be ensured that in each batch, each arm is allocated at least once, imposing an additional restriction on the TS algorithm.
 
 \item[Restricted TS (AW-AIPW)] Allocation probabilities of the TS are restricted to avoid a very fast decreasing or increasing rate towards the extremes $0,1$. Their redefinition is based on a time-varying function~\citep[see][]{hadad2021confidence}, that imposes the lower bound $(1/K) t^{-0.7}$, with $K$ the number of arms, leading to:
 \begin{align*}
    \pi_{\text{min}} &=  \frac{1}{K} t^{-0.7}; \quad \pi_{\text{max}} = 1- \pi_{\text{min}}\\
     \pi^{\text{TS-AWAIPW}}_{t,k} &= \min \left(\pi_{\text{max}},  \max \left(\tilde{\pi}^{\text{TS}}_{t,k}, \pi_{\text{min}}\right)\right), \quad \forall t \in [0,T], k=0,1.     
 \end{align*}
 
\end{description}

\newpage

\section*{D. Additional Experiments}

\subsection*{D1. Sensitivity to non-stationarity}
We report details by focusing on a small-batched setting, i.e., a batch size $n=3$, and again on a sample size $N=150$, as in our real-field experiment illustrated in Section 5 of the main paper.

\begin{figure}[h!]
    \centering
    \includegraphics[scale = .5]{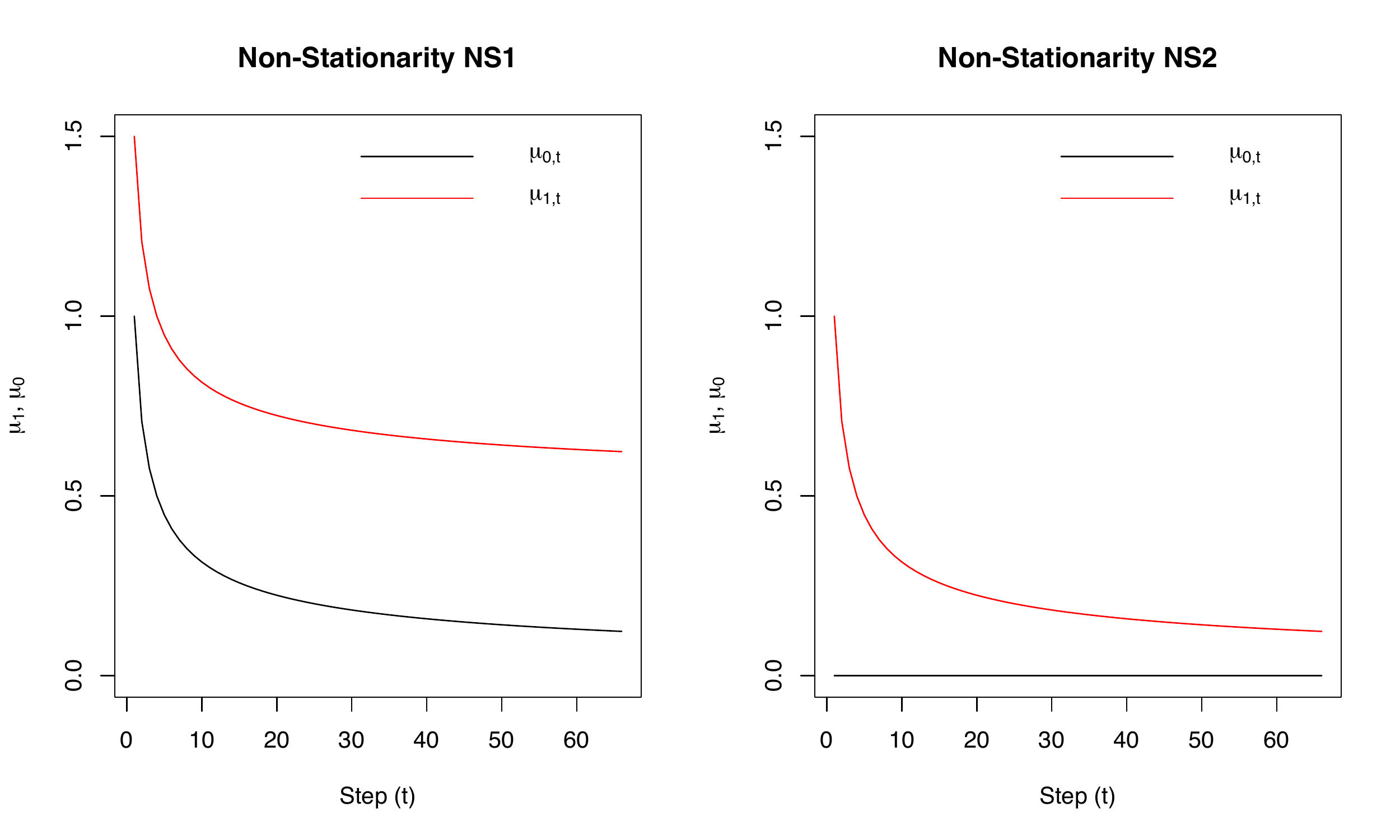}
    \caption{Two types on non-stationarity configuration under the alternative hypothesis $H_1$.}
    \label{fig:ns_H1}
\end{figure}

We hypothesize two types of non-stationarity:
\begin{description}
    \item[NS1] A setting where the arm means vary from one step to another according to a polynomial decay rate, meaning that there's a decreasing time trend in the mean reward, but their difference (i.e., the \textit{treatment effect}, say $\Delta$) remain constant. More specifically, denoting with $\mu$ the baseline (step $t=0$) mean reward, and with $c \in [0,1]$ the decay parameter, we consider the following models under each of the hypothesis:
    \begin{align*} 
    H_0\!&: Y_{t,i} (A_{t,i} = k) \sim \mathcal{N}(\mu_{t,k} = \frac{\mu}{(t+1)^c}, \sigma^2_y),\quad k=0,1\\
    H_1\!&: Y_{t,i} (A_{t,i} = 0) \sim \mathcal{N}(\mu_{t,0} = \frac{\mu}{(t+1)^c}, \sigma^2_y); \quad Y_{t,i} (A_{t,i} = 1) \sim \mathcal{N}(\mu_{t,1} = \mu_{t,0} + \Delta, \sigma^2_y)
    \end{align*}
    
    \item[NS2] A setting where the arm means vary from one step to another according to a polynomial decay rate, and under the alternative the difference between the arm means changes over time as well. More specifically, we consider the following models under each of the hypotheses:
    \begin{align*} 
    H_0\!&: Y_{t,i} (A_{t,i} = k) \sim \mathcal{N}(\mu_{t,k} = \frac{\mu}{(t+1)^c}, \sigma^2_y),\quad k=0,1\\
    H_1\!&: Y_{t,i} (A_{t,i} = 0) \sim \mathcal{N}(\mu_{t,0} = 0, \sigma^2_y); \quad Y_{t,i} (A_{t,i} = 1) \sim \mathcal{N}(\mu_{t,1} = \frac{\mu}{(t+1)^c}, \sigma^2_y)
    \end{align*}    
\end{description}
In both non-stationary cases, we assume $\sigma^2_y = 10$, $\mu=1$ and $c=0.5$, and in the NS1 case, we consider a $\Delta = 0.5$ (similarly to the main paper's experiments). An illustration of the (non-stationary) time trend of the mean rewards is reported in Figure~\ref{fig:ns_H1}.

\begin{figure}[h!]
    \centering
    \includegraphics[scale = .55]{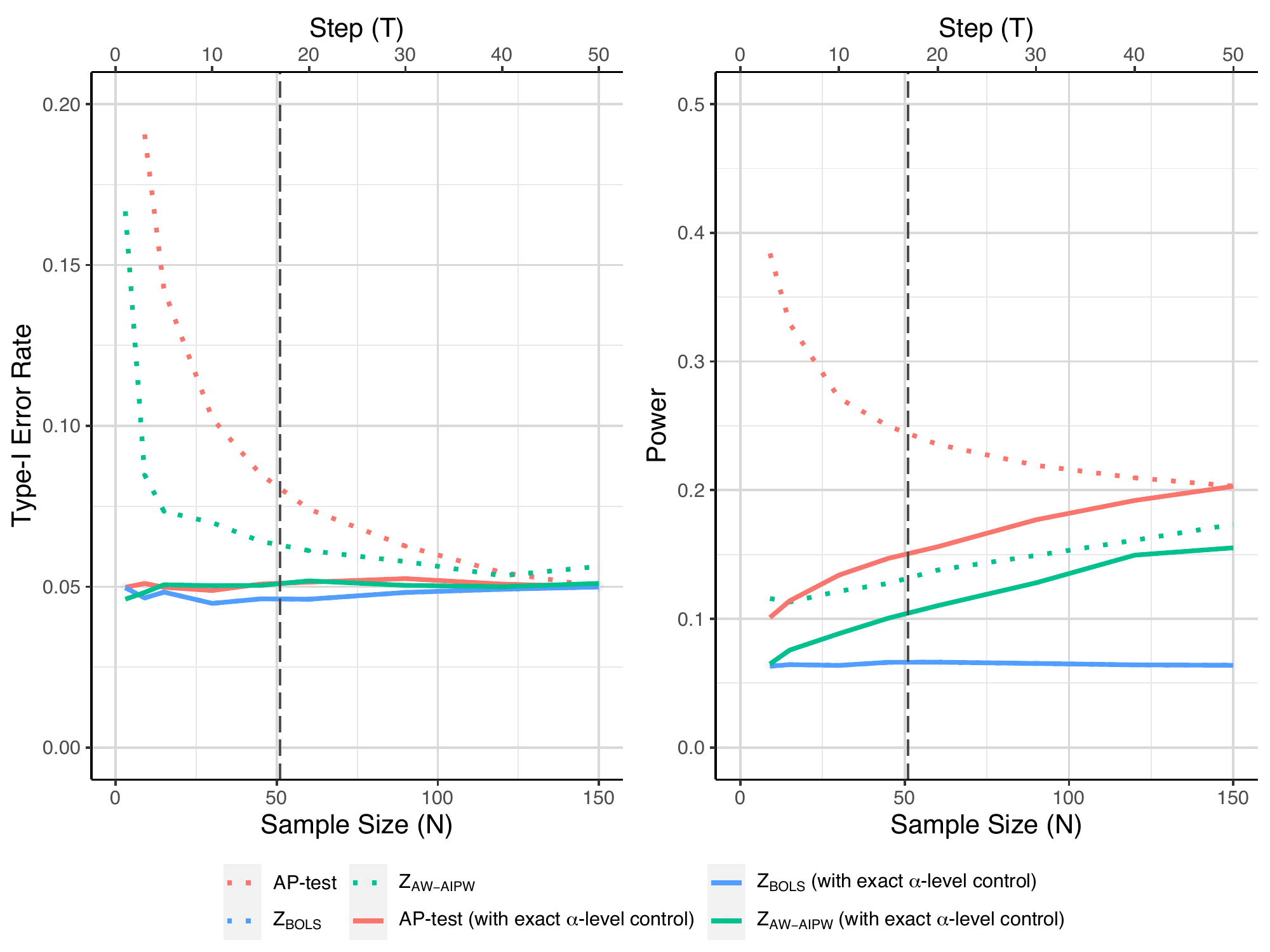}
    \caption{Type-I error and power of the compared approaches in a two-arm setting with $N=150$, $T=50$ and $n=3$. Reward data are simulated according to the non-stationary NS1 case. All values are obtained by averaging across $10^4$ independent trials generated with Thompson sampling, with standard errors always $< 0.005$.}
    \label{fig:typeIer_power_ns1}
\end{figure}

\begin{figure}[h!]
    \centering
    \includegraphics[scale = .55]{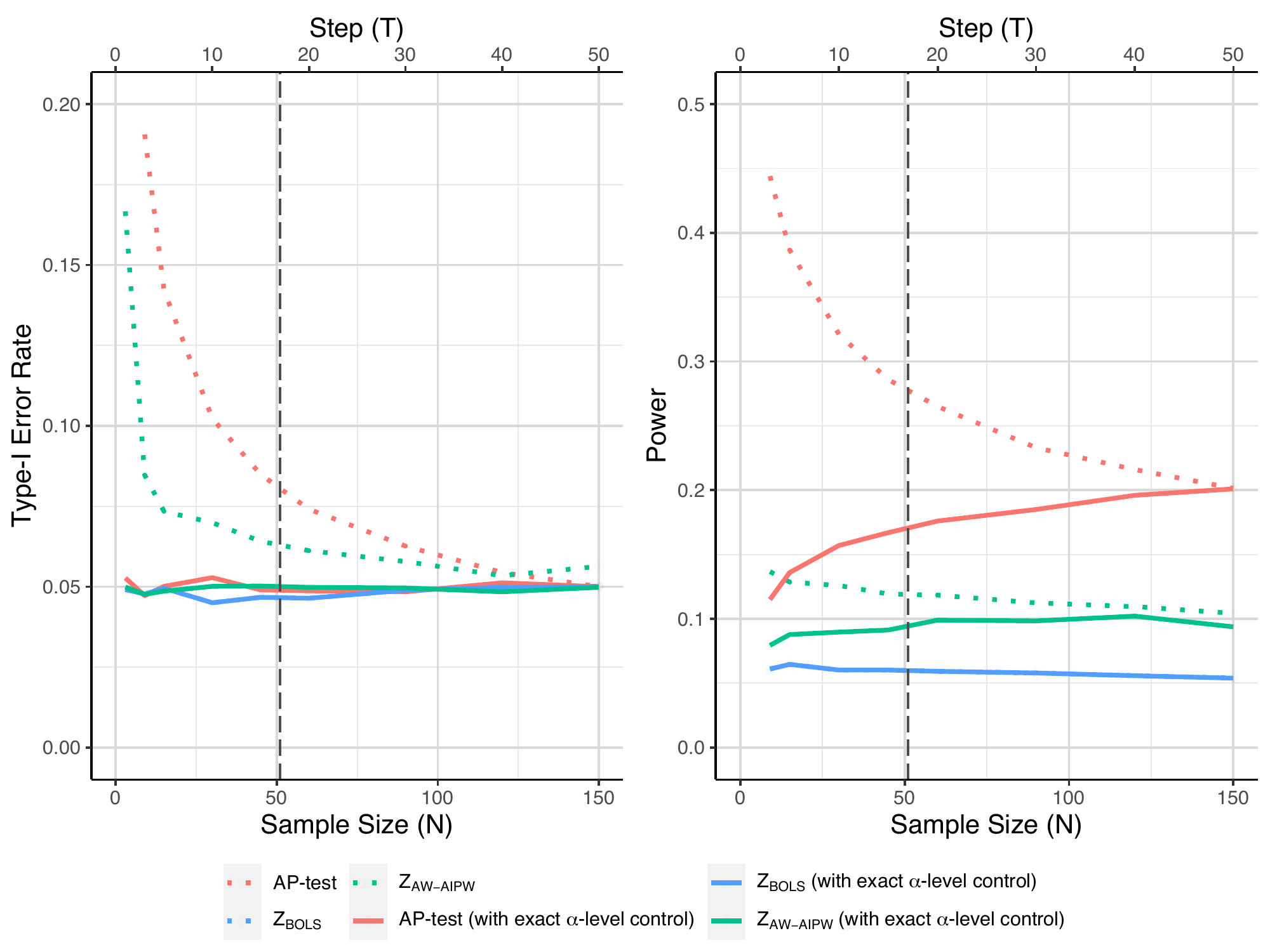}
    \caption{Type-I error and power of the compared approaches in a two-arm setting with $N=150$, $T=50$ and $n=3$. Reward data are simulated according to the non-stationary NS2 case. All values are obtained by averaging across $10^4$ independent trials generated with Thompson sampling, with standard errors always $< 0.005$.}
    \label{fig:typeIer_power_ns2}
\end{figure}

In Figure~\ref{fig:typeIer_power_ns1} and \ref{fig:typeIer_power_ns2} we show the performance of the compared methods in these two non-stationary settings. The first notable result is the overall good sensitivity performances of the proposed AP-test to non-stationarities. Both type-I error and power are consistent among different non-stationarity settings and in comparison with the stationary case. A similar consistency is also shown by the BOLS procedure, which is however very inefficient (i.e., almost no power) as in the stationary case, with a slightly worse performance in the NS2 setting. A different pattern characterizes the AW-AIPW strategy, which, while maintaining an inferential performance similar to the stationarity case in the NS1 type of non-stationarity (see Figure~\ref{fig:typeIer_power_ns1}), it registers a substantial loss in terms of power in the NS2 setting (see Figure~\ref{fig:typeIer_power_ns2}).

\subsection*{D2. Type-I Error and Power in higher batch sizes.} 
\begin{figure}[h!]
    \centering
    \includegraphics[scale = .8]{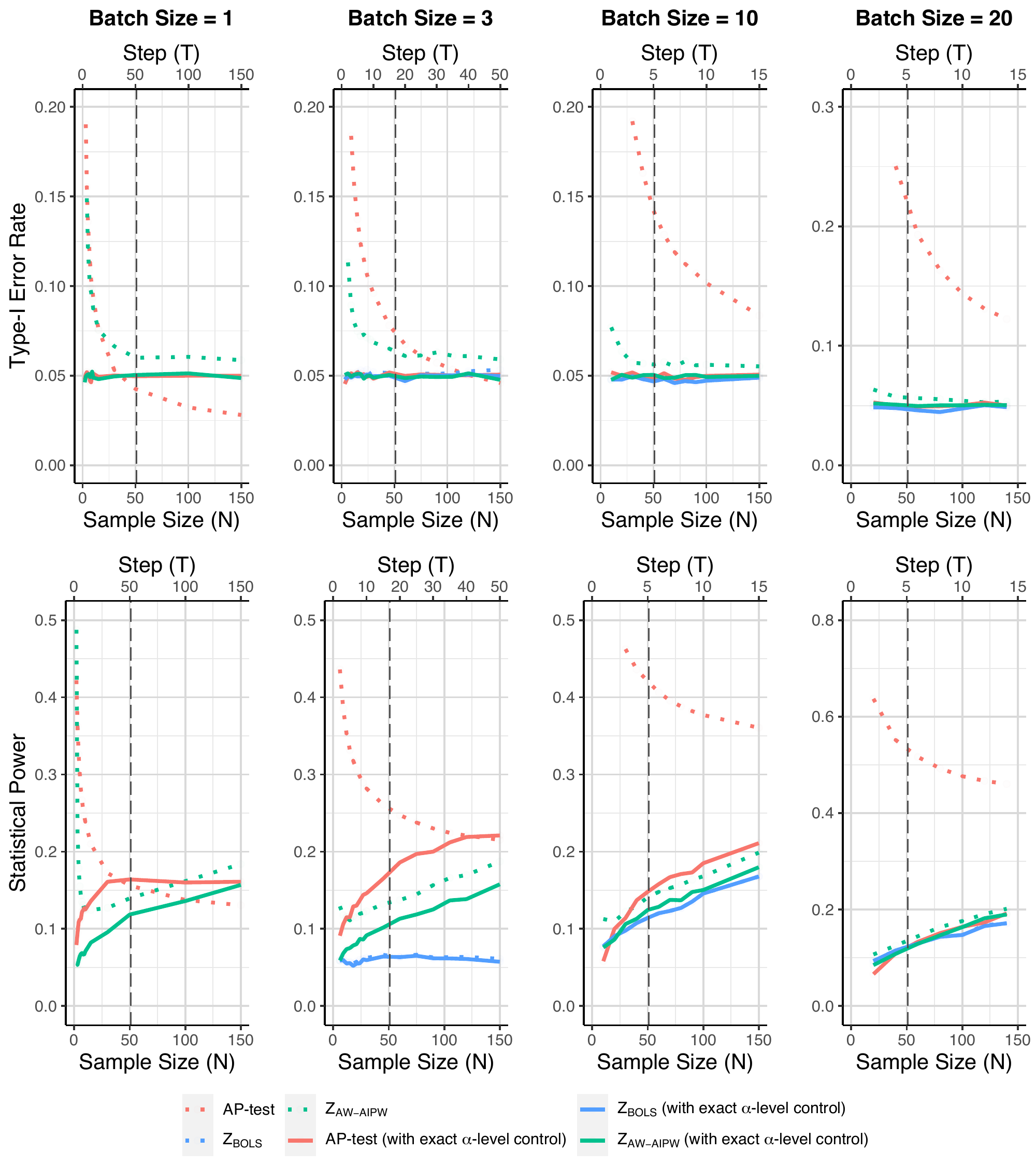}
    \caption{Type-I error and power of the compared tests for different batch sizes, in a two-arm setting under $H_0\!: \mu_0 =  \mu_1$  and $H_1\!: \mu_1 > \mu_0$. Reward data generated according to $Y_{t,i} (A_{t,i} = k) \sim N(0, \sigma^2 = 10)$, $k=0,1$ (for $H_0$), and $Y_{t,i} (A_{t,i} = 0) \sim N(0, \sigma^2 = 10)$ and $Y_{t,i} (A_{t,i} = 1) \sim N(0.5, \sigma^2 = 10)$, $t=0,\dots,T$, $i=1,\dots,n$ (for $H_1$). All values are obtained by averaging across $10^4$ independent trials generated with Thompson sampling.}
    \label{fig:typeIer_power}
\end{figure}

In Figure~\ref{fig:typeIer_power} we show type-I error and power of the proposed AP-test and the comparative approaches, for a higher number of batch sizes. Generally, the BOLS-based asymptotic test shows better type-I error control. This is also because a simulation-based approach is used for estimating its distribution and critical value (as discussed in Section C1). However, its power is extremely low, especially for small sample sizes. 
Increasing the time steps, type-I error remains controlled with the BOLS-based Z-test, and continuously improves with the AW-AIPW-based approach and the proposed AP-test.
As we can see in Figure~\ref{fig:typeIer_power}, in general, a critical value of $T-1$, ensures a nominal $\alpha=0.05$ type-I error control when $t\geq 40$, regardless the batch size. This is not true for the AW-AIPW-based Z-test, which keeps a slightly inflated type-I error with $t$.

In terms of batch sizes, we can see that higher the batch size, better the performances of the BOLS approach, whose theoretical properties hold for $n \to \infty$. For example, in a (high) batch size of $n=20$, its performances are comparable to those of the other procedures.

\section*{E. Extensions to a General $(K+1)$ Multi-Armed Bandit Setting}

\subsection*{E1. General Setup}
We start again by considering $T$ time steps in which individuals are accrued in batches of pre-specified and fixed size $n_t = n \: \forall t$, leading to a total experiment sample size $N = T \times n$. In a general, say $(K+1)$-armed bandit setting, at each step $t$, to each participant $i=1,\dots,n$ we assign an arm $\{A_{t,i}\} \in \{0,\dots,K\}$, where with $0$ we typically label the \textit{control} or ``do-nothing'' arm. Arms are drawn according to a policy $\pi_t \doteq \{\pi_{t,k}, k=0, \dots, K \}$, where each $\pi_{t,k}$ is the time-varying allocation probability of arm $k$ at step $t$. 

According to the potential outcomes framework~\cite{neyman1923,rubin1974estimating,robins1986new}, we define $Y_{t,i}(A_{t,i})$ to be the random reward variable representing the outcome that would be observed if participant $i$ at step $t$ were assigned to arm $A_{t,i}$. Note that each participant can be assigned only one arm, thus, we observe only one realized outcome for that participant. Denoted by $\mathcal{H}_{t-1} \doteq \{A_{\tau,i}, Y_{\tau,i}(A_{\tau,i}), i=1,\dots, n, \tau = 1,\dots,t-1 \}$, the history of selected arms and associated rewards prior to current step $t$, the sampling probabilities are a function of this history, i.e., $\pi_{t,k} \doteq \mathbb{P}(A_{t,i}=k | \mathcal{H}_{t-1})$, for $k = 0,\dots, K$ and $t=0,\dots,T$. We assume that the reward variables are drawn independently from an unknown but fixed distribution with the following conditional mean:
\begin{align}\label{eq:reward_model_k}
    \mathbb{E}(Y_{t,i}(A_{t,i})| \mathcal{H}_{t-1}) = \mathbb{E}(Y_{t,i}| \mathcal{H}_{t-1}, A_{t,i}) =  \sum_{k=0}^{K}\mu_{t,k}\mathbb{I}(A_{t,i}=k),
\end{align}
where $\mathbb{I}(A_{t,i}=k)$, for $k = 0,\dots, K$, is the identity function assuming value $1$ when arm $k$ is allocated to individual $i$ at step $t$, and $0$ otherwise, and $\{\mu_{t,k}, k=0,\dots,K, t=0,\dots, T \}$ are the unknown arm parameters, or \textit{treatment effects}. As one can notice, here, we also relax the stationarity assumption, allowing the conditional reward mean to vary across steps, that is, we allow the unknown arm parameters $\{\mu_{t,k}, k=0,\dots,K, t=0,\dots, T \}$ to vary with respect to $t$ as well.

\subsection*{E2. Multiple Hypothesis Testing Problem}
In a multi-arm setting, a hypothesis testing problem needs to be formulated in accordance with experimenter interest and the potentially multiple arms to be compared. If the interest remains in comparing a single experimental arm with a single control arm of interest, then the setting is equivalent to the one discussed in the two-arm setting. However, if the interest is in comparing an experimental arm of interest with all the other trial arms, the hypothesis testing involves simultaneous testing of more than one hypothesis (i.e, \textit{multiple testing}).  In such a case, standard procedures typically involve some kind of adjustment on the significance level $\alpha$. Indeed, if decisions about the individual hypotheses are based on an unadjusted value, then there is typically a large probability that some of the true null hypotheses will be rejected.

Focusing again on a superiority, but multiple, hypothesis testing, and denoting with $k^*$ the superior experimental arm, we specify our \textit{null hypotheses} $H_{0,1},\dots, H_{0,K}$ and the \textit{alternative} hypothesis $H_1$ as:
\begin{align}
    &H_{0,k}: \mu_{k^*} = \mu_{k},\quad  k \neq k^*\quad \text{there is no difference between the experimental and the other arms},\label{eq:H0}\\
    &H_1: \mu_{k^*} > \mu_{k},\quad \forall k \neq k^*  \quad \text{the experimental arm } k^* \text{is superior to all the other arms.}\label{eq:H1}
\end{align}

Note that both the null and the alternative hypotheses are specified according to the research interest. Here, we are interested in the superiority of the experimental arm compared to \textbf{all} the other (control) arms; in such a case, the null hypothesis involves testing a number of $K$ hypothesis, equivalent to the number or comparisons. However, other interests may be related to a difference between the experimental and at least one other (control) arm of the experiment.

A general attempt at inference is to naively perform a classical hypothesis test following the procedure illustrated in Section 2.2 (main paper), i.e., controlling the probability of a type-I error for each member of the family of hypotheses. More formally, given the null hypotheses specified in Eq.~\eqref{eq:H0}, a generic test statistic $T(Y)$, and a significance level $\alpha$, one may define the $K$'s $t_{\alpha, k}$ critical values as:
\begin{align}\label{eq:typeIer}
    t_{\alpha, k}:\ \mathbb{P}\left(\text{Reject}\ H_{0,k} | H_{0,k}\ \text{is true}\right) = \mathbb{P}\left(T(Y)>t_{\alpha,k} | \mu_{k^*} = \mu_{k} \right) = \alpha,\quad \forall k \neq k^*, 
\end{align}
and testing each hypotheses as in a single null hypothesis testing problem. 

However, if this naive approach is adopted, there is typically a large probability that some of the true null hypotheses will be rejected. Thus, for identifying the \textit{critical value} of a test statistic $T(Y)$, and the \textit{rejection} or \textit{critical region}, based on which the null hypotheses are rejected, one must consider an adapted version of the type-I error. 
In statistics, the ``global'' type-I error when performing multiple hypotheses tests, is known as family-wise error rate (FWER), and it represents the probability of making one or more false rejections or discoveries, i.e., 
\begin{align*}
    \mathrm{FWER} &= \mathbb{P}(\text{Reject at least one of the $K$ null hypothesis}\ |\ \text{all}\ \{H_{0,k}, k \neq k^*\}\ \text{are true}) \\ &=\mathbb{P}(\text{At least one type-I error on $K$ tests}) = \mathbb{P}(V \geq 1) = 1 - \mathbb{P}(V = 0),
\end{align*}
where $V$ is the number of type-I errors (also called \textit{false positives} or \textit{false discoveries}). Thus, by assuring $\mathrm{FWER} \leq \alpha$, the probability of making one or more type-I errors in the family is controlled at level $\alpha$.

Controls for multiplicity are methods for controlling the rate at which type-I errors occur when conducting multiple hypotheses tests simultaneously. These are called \textit{simultaneous testing procedures} and involve adjusting either the relevant test statistics or critical value. One proposal for achieving such a control is given by \textit{Sidak}'s multiple testing procedure, based on testing each hypothesis at level $\alpha_{SID}=1-(1-\alpha )^{\frac{1}{m}}$, where $m$ refers to the number of null hypotheses to test, in our case $m = K$.

Analogously to type-I error, an adjustment is also required for the definition of the type-II error, or alternatively power. Depending on the objective of the experiment, the power of the study can be defined in different ways. Bretz and colleagues~\citep{bretz2016multiple,senn2007power,vickerstaff2019methods} describe the common underlying theory of multiple comparison procedures through numerous examples, and discuss the difference between: i) ``disjunctive power'' (or minimal power), ii) ``conjunctive power'' (or maximal power) and iii) ``marginal power''. The conjunctive power is the one associated with our definition of the alternative hypothesis in Eq.~\eqref{eq:H1}, i.e., the probability of finding a true treatment difference (more specifically, superiority) when comparing the experimental and \textbf{all} the other (control) arms:
\begin{align*}
    \mathrm{Power} = \mathbb{P}(\text{Reject all the $K$ null hypothesis} | H_1\ \text{is true}). 
\end{align*}

While multiplicity adjustments protect against spurious rejections when there are multiple statistical tests, an important consequence is a change in statistical power. It is typically argued that multiplicity adjustments result in a loss of power, which can be substantial. Thus, an adequate adjustment of a power analysis (or sample size calculation) is required; see e.g.,~\cite{vickerstaff2019methods} and \cite{chow2007sample} for a discussion around this issue.

\subsection*{E3. Thompson Sampling Allocation Probability Test} In a general $(K+1)$-arm setting, the Thompson sampling approach defines the allocation probability of a given arm, say $k$, as the posterior probability of that arm being associated with the maximum expected reward, i.e.,
\begin{align}\label{eq:TS_alloprob}
    \pi_{t,k}^{\text{TS}} = \mathbb{P}\Big(\mathbb{E}\big(Y_{t,i}(A_{t,i} = k)\big) > \mathbb{E}\big(Y_{t,i}(A_{t,i} = k')\big) | \mathcal{H}_{t-1} \Big), \quad \forall k' \neq k.
\end{align} 

The common \textit{posterior sampling} idea of TS, involves, first, drawing at the beginning of each step $t$ a sample from the posterior distribution of each of the unknown treatment effect parameters, say $\{\tilde{\mu}_{t,k}, k=0,\dots,K\}$. Then, for each arm, computing the posterior estimated mean reward $\mathbb{E}(\tilde{Y}_{t,i}|A_{t,i} = k) = \tilde{\mu}_{t,k}$, and selecting the arm $\tilde{a}_{t,i}$ associated with the highest mean reward, i.e.,
\begin{align}\label{eq:TS_atilde}
    \tilde{a}_{t,i} \doteq \text{argmax}_{k = 1,\dots,K}\mathbb{E}(\tilde{Y}_{t,i}|A_{t,i} = k) = \text{argmax}_{k = 1,\dots,K}\tilde{\mu}_{t,k}.
\end{align}

As with an increased number of arms the exact computation of the allocation probabilities as in Eq.~\eqref{eq: exact_TS} becomes intensive, \textit{Monte Carlo} (MC) simulations can be used for calculating the allocation probability of each arm $k$ and step $t$: this is given by the proportion of times (over the MC runs) that arm $k$ was selected by satisfied the maximizing condition in Eq.~\eqref{eq:TS_atilde}. More formally, assuming $M$-MC runs, the estimated allocation probability for each arm $k$ and at each step $t$ is calculated as:
\begin{align*}\label{eq:TS_est_alloprob}
    \hat{\pi}_{t, k}^{\text{TS}} = \frac{\sum_{m = 1}^M \mathbb{I}(\tilde{\mu}_{m,t,k} > \tilde{\mu}_{m,t,k'} | \mathcal{H}_{t-1})}{M}, \quad \forall k' \neq k.
\end{align*}

Given these (exact or estimated) probabilities, more specifically the allocation probability of the experimental arm $k^*$ (as it is the one we are interested in), we define again our AP-test as the number of steps $t \leq T$ in which the allocation probabilities exceeded a fixed value $\pi^{\text{ER}}$ that reflects an equal randomization (ER) design, i.e., 
\begin{align*}
    \text{AP}^\text{TS}_T \doteq \sum_{t=1}^T\mathbb{I}\left(\hat{\pi}_{t,k^*} > \pi^{\text{ER}} \right),
\end{align*}
with $k^*$ the experimental arm.
Note that, in a general $(K+1)$-arm setting, the fixed threshold is given by the reciprocal of the number of arms, i.e., $\pi^{\text{ER}} \doteq 1/(K+1)$. The idea is that, higher the value of the test statistic, i.e., higher the number of steps in which the allocation probability to the experimental arm exceeded the ER threshold $\pi^{\text{ER}}$, higher the evidence that the experimental arm is superior. 

The exact distribution under certain bandit strategies, such as TS, could be computed following the logic illustrated in the previous Section A. Computation of the critical value and the related type-I error and power follows from the definition given in Eq. (6) in the main paper.

Finally, an extension of Theorem 3.2 and Lemma 3.2 is provided as follows.

\begin{customthm}{Theorem 3.3}
Assuming that the experimental arm is superior to the control arms, i.e., $\mu_{k^*} > \mu_k$, for all $k^* \neq k$ and that arm's allocation probabilities are defined according to standard Thompson Sampling policy, then the allocation probability AP-test diverges as $T \to \infty$, i.e., $\text{AP}_T \to \infty$.
\end{customthm}

\begin{customthm}{Lemma 3.4}
Assuming that there is a unique optimal arm, with $\mu_{k^*} > \mu_k$, for all $k^* \neq k$, as the arms are allocated an arbitrarily large number of times, TS converges to allocating the arm $k^*$ with probability $1$ as $t \to \infty$, i.e.,
\begin{align*}
    \lim_{t \to \infty} \pi_{t,k^*}^{TS} = 1,\quad \text{when}\  \mu_{k^*} > \mu_k \quad \forall k^* \neq k.
\end{align*}
\end{customthm}
Proofs follow the same procedure adopted in Section A.

\section*{F. Code and Data for Results Reproducibility}
\textbf{Code.} All the codes used to the generate the simulated data and to perform inference will be provided with the camera-ready version.

\textbf{Data.} Real-world experiments data, including both the Thompson Sampling based and the fixed ER deployment (used as a benchmark for ``validating'' TS outcomes), will be made available with the camera-ready version.

\end{document}